\documentclass[10pt,twocolumn,letterpaper]{article}

\usepackage{cvpr}              
\usepackage{amsmath}
\usepackage{algorithm}
\usepackage{algpseudocode}
\usepackage{amsfonts}
\usepackage{color}
\usepackage{graphicx} 
\usepackage{array} 
\usepackage{xcolor,colortbl} 
\definecolor{lightgray}{gray}{.9}
\definecolor{forestgreen}{RGB}{0,128,69}
\definecolor{cvprpink}{RGB}{219,112,147}
\definecolor{cvpryellow}{RGB}{255, 191, 0}
\usepackage{booktabs} 
\usepackage{multirow} 
\usepackage{multicol} 
\usepackage{caption} 

\usepackage{mathtools}

\usepackage{bm}

\usepackage[export]{adjustbox}

\usepackage{threeparttable}
\newcolumntype{I}{!{\vrule width 1pt}}
\makeatletter
\newcommand{\thickhline}{%
    \noalign {\ifnum 0=`}\fi \hrule height 1pt
    \futurelet \reserved@a \@xhline
}
\makeatother
\usepackage{float}
\usepackage{pifont}
\usepackage{ragged2e}
\usepackage{enumitem}

\newtheorem{definition}{Definition}

\definecolor{cvprblue}{rgb}{0.21,0.49,0.74}

\usepackage[pagebackref,breaklinks,colorlinks]{hyperref}

\def\paperID{2420} 
\def\confName{CVPR}
\def\confYear{2025}

\title{Geometric Knowledge-Guided Localized Global Distribution Alignment for Federated Learning}

\author{
    Yanbiao Ma\thanks{Co-first authors}\\
    Xidian University\\
    {\tt\small ybmamail@stu.xidian.edu.cn}
    \and
    Wei Dai\footnotemark[1]\\
    Xidian University\\
    {\tt\small 22012100039@stu.xidian.edu.cn}
    \and
    Wenke Huang\thanks{Corresponding Author: Wenke Huang}\\
    Wuhan University\\
    {\tt\small wenkehuang@whu.edu.cn}
    \and
    Jiayi Chen\\
    Xidian University\\
    {\tt\small 22012100031@stu.xidian.edu.cn}
}

\begin{document}
\maketitle
\vspace{-20pt}
\begin{abstract}
Data heterogeneity in federated learning, characterized by a significant misalignment between local and global distributions, leads to divergent local optimization directions and hinders global model training. Existing studies mainly focus on optimizing local updates or global aggregation, but these indirect approaches demonstrate instability when handling highly heterogeneous data distributions, especially in scenarios where label skew and domain skew coexist. To address this, we propose a geometry-guided data generation method that centers on simulating the global embedding distribution locally. We first introduce the concept of the geometric shape of an embedding distribution and then address the challenge of obtaining global geometric shapes under privacy constraints. Subsequently, we propose GGEUR, which leverages global geometric shapes to guide the generation of new samples, enabling a closer approximation to the ideal global distribution. In single-domain scenarios, we augment samples based on global geometric shapes to enhance model generalization; in multi-domain scenarios, we further employ class prototypes to simulate the global distribution across domains. Extensive experimental results demonstrate that our method significantly enhances the performance of existing approaches in handling highly heterogeneous data, including scenarios with label skew, domain skew, and their coexistence. Code published at: \url{https://github.com/WeiDai-David/2025CVPR_GGEUR}
\end{abstract}

\vspace{-10pt}
\section{Introduction}
\label{sec:intro}

\begin{figure}[t]
  \centering
   \includegraphics[width=1\linewidth]{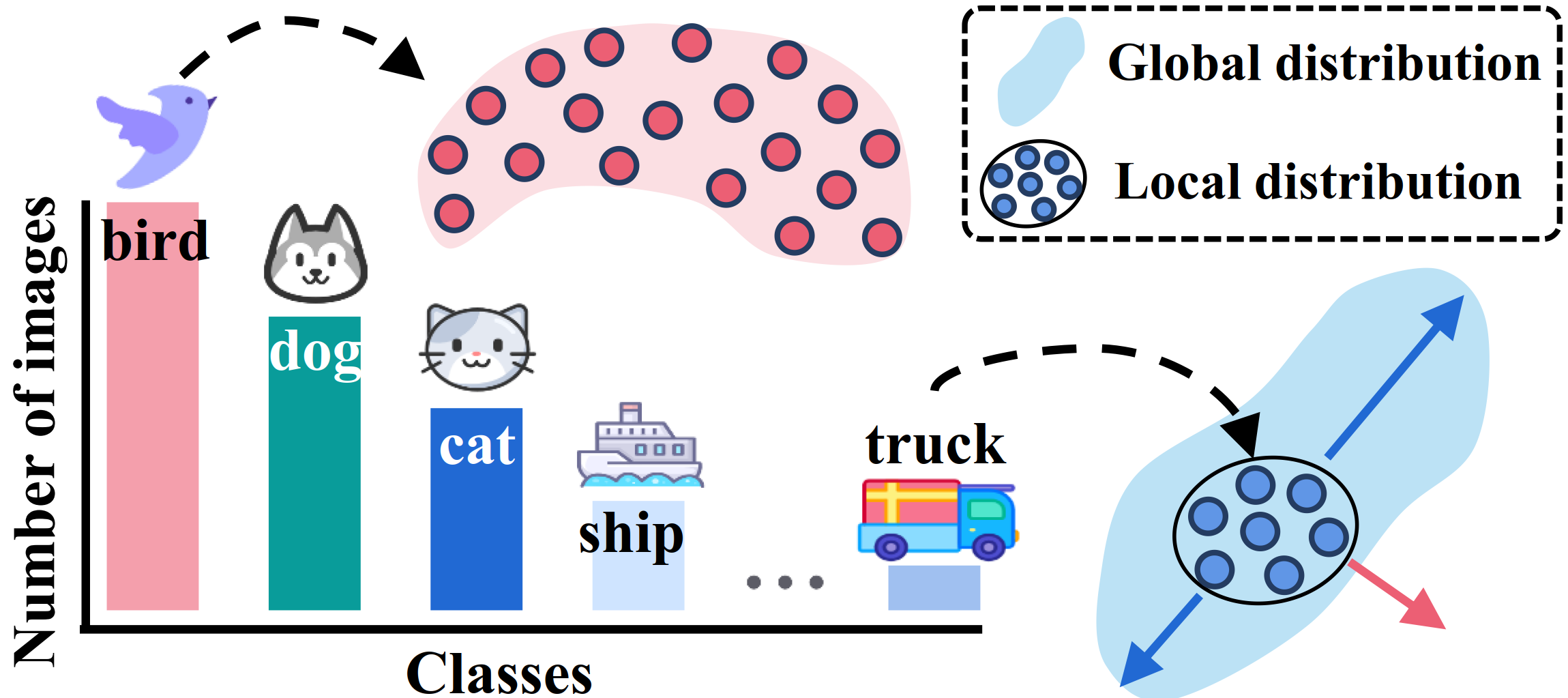}
\vskip -0.05in
   \caption{The distribution of local data on clients is imbalanced. The ``bird" category has a large number of samples, allowing its local distribution to effectively represent the global distribution. In contrast, the ``truck" category has insufficient samples, resulting in a significant disparity between the local and global distributions.}
   \label{fig1}
\vskip -0.15in
\end{figure}

Federated Learning (FL) enables multiple clients to collaboratively train a global model without sharing raw data, effectively addressing key concerns related to data privacy and security \cite{fl1,fl2,fl3}. However, data heterogeneity remains one of the primary challenges in FL, especially in the presence of label skew and domain skew across clients \cite{huang3,jiang2022towards,karimireddy2020scaffold,li2020federated,li2}. Such distributional differences lead to variations in the local optimization directions for each client, which can be particularly pronounced when data is sourced from multiple domains \cite{huang2,li2023no,wu2023fediic,zhao2018federated,mu2023fedproc}. This cross-client distribution shift not only slows down global model convergence but also undermines the model’s generalization capability across clients \cite{fl3,huang2,ye2023heterogeneous,li2021fedbn}.

Existing research addressing data heterogeneity in federated learning primarily focuses on two categories of methods \cite{fl3}. The first is server-side optimization strategies, which aim to enhance model convergence by adjusting the global aggregation process \cite{wang2020federated,al2020federated,singh2020model,xia2021auto,sun2023shapleyfl,chen2023elastic,li2023revisiting,palihawadana2022fedsim,huang4,zhang2022fedada}. The second category includes client-side regularization \cite{mu2023fedproc,huang3,acar2021federated,gao2022feddc,li2022partial,lee2022preservation} and data augmentation methods \cite{zhou2023fedfa,zhu2021data}, which attempt to align local model optimization through guidance from the global model and loss function adjustments or directly apply data augmentation techniques like Mixup in the local setting \cite{yoon2021fedmix}. However, these approaches do not directly tackle the core issue of data heterogeneity, namely, the significant misalignment between local data distributions on clients and the global data distribution. They often rely on indirect means to optimize the alignment between local and global models, which typically incurs high computational costs and depends heavily on the selection of auxiliary datasets, resulting in limited stability \cite{fl3}. 
\textbf{Motivated} by these challenges, we propose a core idea: if clients can locally simulate an ideal global distribution, it may be possible to alleviate the divergence in local optimization directions, even without data sharing. In other words, through guided data augmentation, we can reduce the inconsistency between local and global distributions, fundamentally mitigating data heterogeneity issues in federated learning.

How can we simulate a global distribution locally? Taking CIFAR-10 as an example, Figure \ref{fig1} illustrates the label distribution in a random client, revealing severe imbalances. Certain categories, such as ``truck," are underrepresented, making it difficult to form an effective global representation. Given privacy constraints, simulating a global distribution locally becomes a significant challenge \cite{fl3}. As shown in the blue distribution in Figure \ref{fig1}, augmenting samples along the blue arrow direction, rather than the pink arrow, is more likely to simulate the global distribution. Thus, \textbf{identifying appropriate augmentation directions and ranges becomes our key objective to overcome this challenge.}

\begin{figure}[t]
  \centering
   \includegraphics[width=1\linewidth]{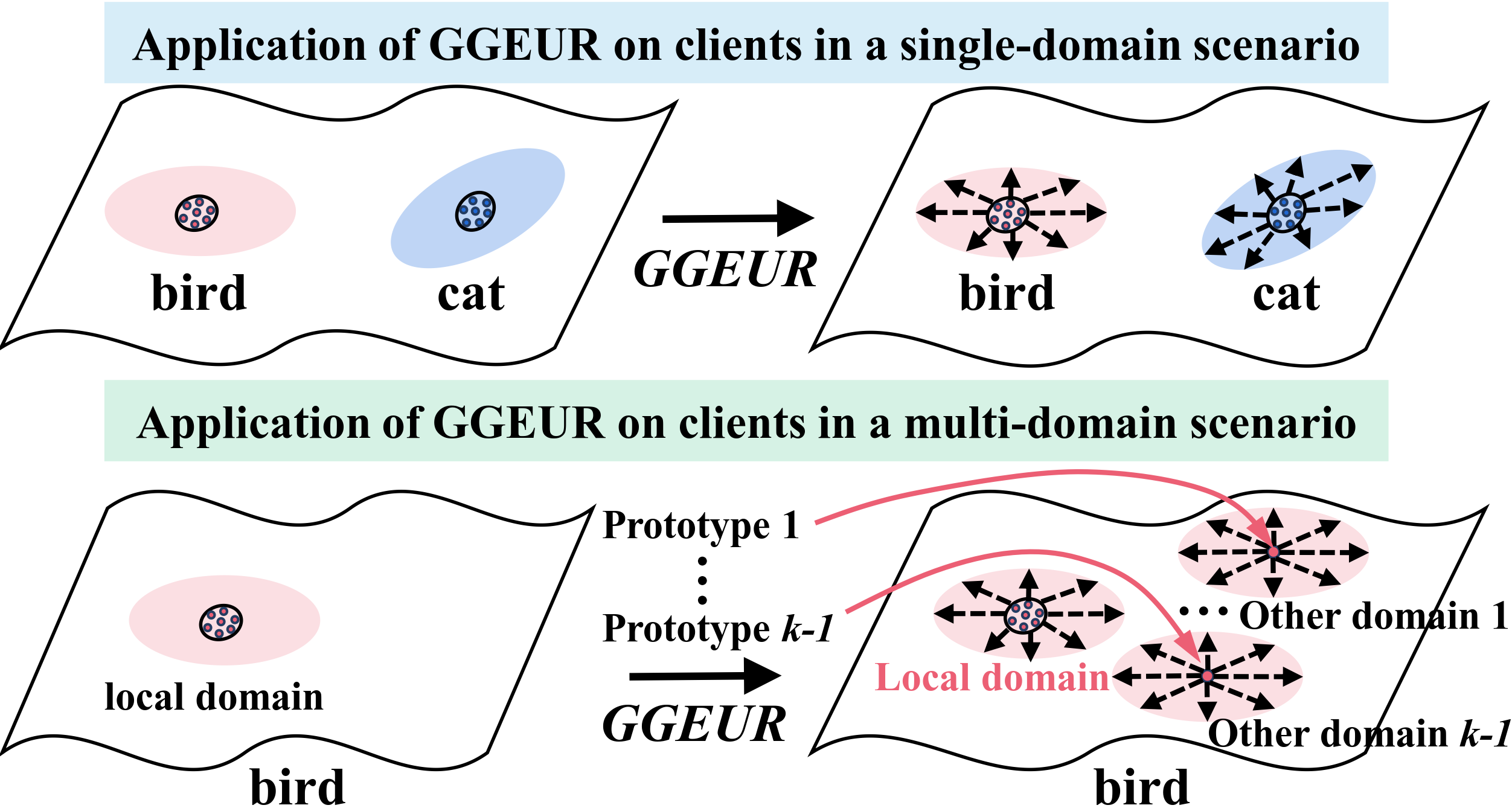}
\vskip -0.05in
   \caption{Example of ``bird" category: in a single-domain scenario, only the local global distribution is simulated; in a multi-domain scenario, the global distribution from other data domains is also simulated on a client.}
   \label{fig2}
\vskip -0.15in
\end{figure}

In the single-domain scenario, some clients may possess a large number of ``truck" images, while others have only a few. The aggregate set of ``truck" images across all clients constitutes the global distribution for that category. \textbf{We introduce} the concept of geometric shapes to describe the orientation and range of data distributions (see Section \ref{sec2}). For instance, the global distribution of ``truck" in Figure \ref{fig1} spans primarily from the bottom left to the top right. If we could quantify the global geometric shape of the "truck" and disseminate it to clients, they could use it to guide new sample generation locally. However, privacy constraints prohibit us from aggregating all client samples to quantify this global geometric shape. To address this, \textbf{we propose} using local covariance matrices from multiple clients on the server side to approximate the global covariance matrix, enabling quantification of the global geometric shape (see Section \ref{sec3.1}). To mitigate inaccuracies in local covariance matrix estimation due to high dimensionality and sparsity in image space \cite{ma2023delving}, we employ the CLIP \cite{clip} to extract embedding distributions from local client data and carry out the approximation process in the embedding space.

After obtaining the global geometric shape for each category, \textbf{we propose} a method called Global Geometry-Guided Embedding Uncertainty Representation (\textbf{GGEUR}) to generate new samples on clients, simulating an ideal global distribution in the embedding space. Each client then only needs to train an MLP as a local classifier on the augmented data. In multi-domain scenarios, we observe that the geometric shapes of the embedding distributions for the same category across different domains exhibit similarity (see Figure \ref{fig6} and Section \ref{sec3.2}). This makes it feasible to approximate the global geometric shape using local geometric shapes across all clients, even in multi-domain settings. In multi-domain scenarios, however, the global distribution for each category comprises distributions from all domains, necessitating that clients also simulate embeddings from other domains. Although the embedding distributions of the same category across different domains are geometrically similar, they occupy different spatial positions. Thus, unlike the single-domain scenario, we propose distributing category prototypes from other domains as shared knowledge to clients and applying GGEUR to these prototypes to generate new samples that simulate distributions from other domains (see Section \ref{sec3.2}). Figure \ref{fig2} illustrates the difference between single-domain and multi-domain scenarios.

Our approach, serving as a preprocessing step in FL, is easily integrated with other methods. Extensive evaluation across scenarios with label skew, domain skew, and combined skew demonstrates that our method significantly improves the performance of existing approaches, achieving state-of-the-art results in highly heterogeneous data scenarios (Section \ref{sec4}). This work represents a typical instance of synergy between foundational visual models and geometric knowledge within FL.

\section{Geometry of Embedding Distributions}
\label{sec2}

We now define the geometry of an embedding distribution and introduce a measure of similarity between these geometries. In a $P$-dimensional space, given a dataset of a certain class $X = [x_1, \dots, x_n] \in \mathbb{R}^{P \times n}$, the covariance matrix of the distribution can be estimated as:
\begin{equation}
\begin{split}
\Sigma_X=\mathbb{E}\left[\frac{1}{n} {\textstyle \sum_{i=1}^{n}x_ix_i^T}\right ]=\frac{1}{n}XX^T\in \mathbb{R}^{P\times P}.
\end{split}
\end{equation}
Performing eigenvalue decomposition on $\Sigma_X$ yields $P$ eigenvalues $\lambda_1 \geq \cdots \geq \lambda_P$ and their corresponding $P$-dimensional eigenvectors $\xi_1, \dots, \xi_P$. All eigenvectors are mutually orthogonal, collectively anchoring the skeleton of the data distribution, with each eigenvalue specifying the range of the distribution along the direction of its corresponding eigenvector. The combination of eigenvalues and eigenvectors defines the geometric shape of the distribution. Thus, we define the geometric shape of the data $X$ as:
\begin{equation}
\setlength\abovedisplayskip{2pt} 
\setlength\belowdisplayskip{2pt}
\begin{split}
GD_X = \{\xi_1, \dots, \xi_P, \lambda_1, \dots, \lambda_P\}.
\end{split}
\end{equation}

Given the geometries $\text{GD}_{X_1}$ and $\text{GD}_{X_2}$ of two distributions, their similarity is defined as:
\begin{equation}
\setlength\abovedisplayskip{2pt} 
\setlength\belowdisplayskip{2pt}
\begin{split}
S(GD_{X_1},GD_{X_2})=\sum_{i=1}^{P}\left | \left \langle \xi_{X_1}^i,\xi_{X_2}^i \right \rangle \right |.
\end{split}
\end{equation}
where $S(\text{GD}_{X_1}, \text{GD}_{X_2})$ ranges from $0$ to $P$. A higher value indicates greater similarity in geometry. In this study, we calculate the geometric similarity using the eigenvectors corresponding to the top five eigenvalues.

\section{Simulating the Global Distribution Locally}
\label{sec3}

In this section, we address a \textbf{critical issue}: how to approximate the geometric shape of the global data distribution using local client data while preserving privacy. We then detail how to simulate the ideal global distribution locally in both single-domain and multi-domain scenarios by leveraging the geometric shape of the global distribution.

\subsection{Computation of Global Geometric Shapes Under Privacy Constraints}
\label{sec3.1}

Consider a classification task with \( K \) clients and \( C \) classes. Suppose each client contains \( n_1^i, n_2^i, \dots, n_K^i \) samples belonging to class \( i \), represented by the sample set \( \{ x_k^1, x_k^2, \dots, x_k^{n_k^i} \} \) for client \( k \). The global distribution for class \( i \) is formed by the combined distributions of these local samples across the \( K \) clients. However, we cannot directly access the global distribution to derive its geometric shape. Recalling Section 2, the covariance matrix of a distribution is fundamental for obtaining its geometric shape. Thus, our goal is to approximate the global data distribution's covariance matrix without sharing data.

We propose to approximate the global covariance matrix by leveraging the covariance matrices of all local data. Taking class \( i \) as an example, we first compute the local covariance matrix \( \Sigma_k^i \) and local mean \( \mu_k^i (k=1,\dots,K) \) for class \( i \) on each client as follows:
\begin{small}
\begin{equation}
\begin{split}
\mu_k^i = \frac{1}{n_k^i} \sum_{j=1}^{n_k^i} x_k^{i,j}, 
\Sigma_k^i = \frac{1}{n_k^i} \sum_{j=1}^{n_k^i} (x_k^{i,j} - \mu_k^i)(x_k^{i,j} - \mu_k^i)^T.
\nonumber
\end{split}
\end{equation}
\end{small}
The global covariance matrix is the covariance of the combined data from all clients, defined as:
\begin{equation}
\setlength\abovedisplayskip{2pt} 
\setlength\belowdisplayskip{2pt}
\begin{split}
\mu_i = \frac{1}{N_i} \sum_{k=1}^K \sum_{j=1}^{n_k^i} x_k^{i,j} = \frac{1}{N_i} \sum_{k=1}^K n_k^i \mu_k^i, \\
\Sigma_i = \frac{1}{N_i} \sum_{k=1}^K \sum_{j=1}^{n_k^i} (x_k^{i,j} - \mu_i)(x_k^{i,j} - \mu_i)^T,
\nonumber
\end{split}
\end{equation}
where \( N_i = \sum_{k=1}^K n_k^i \) represents the total number of samples in class \( i \) across all clients. We can decompose \( (x_k^{i,j} - \mu_i) \) as \( (x_k^{i,j} - \mu_k^i) + (\mu_k^i - \mu_i) \), so the global covariance matrix \( \Sigma_i \) can be rewritten as:
\begin{small}
\begin{equation}
\setlength\abovedisplayskip{2pt} 
\setlength\belowdisplayskip{2pt}
\begin{split}
\Sigma_i = \frac{1}{N_i} \sum_{k=1}^K \sum_{j=1}^{n_k^i} [ (x_k^{i,j} - \mu_k^i + \mu_k^i - \mu_i)(x_k^{i,j} - \mu_k^i + \mu_k^i - \mu_i)^T ].
\nonumber
\end{split}
\end{equation}
\end{small}
Expanding the above equation yields:
\begin{small}
\begin{equation}
\setlength\abovedisplayskip{2pt} 
\setlength\belowdisplayskip{2pt}
\begin{split}
\Sigma_i = \frac{1}{N_i} \sum_{k=1}^K \sum_{j=1}^{n_k^i} [ (x_k^{i,j} - \mu_k^i)(x_k^{i,j} - \mu_k^i)^T + (x_k^{i,j} - \mu_k^i)(\mu_k^i - \mu_i)^T  \\
+ (\mu_k^i - \mu_i)(x_k^{i,j} - \mu_k^i)^T + (\mu_k^i - \mu_i)(\mu_k^i - \mu_i)^T].
\nonumber
\end{split}
\end{equation}
\end{small}
By the properties of covariance matrices, the first term is the local covariance matrix \( \Sigma_k^i \), and the expected values of the second and third terms are zero. The fourth term can be computed as:
\begin{small}
\begin{equation}
\setlength\abovedisplayskip{2pt} 
\setlength\belowdisplayskip{2pt}
\begin{split}
\sum_{j=1}^{n_k^i} (\mu_k^i - \mu_i)(\mu_k^i - \mu_i)^T = n_k^i (\mu_k^i - \mu_i)(\mu_k^i - \mu_i)^T.
\nonumber
\end{split}
\end{equation}
\end{small}
Thus, the global covariance matrix can be obtained by combining the local covariance matrices and local means as:
\begin{small}
\begin{equation}
\begin{split}
\Sigma_i = \frac{1}{N_i} \left( \sum_{k=1}^K n_k^i \Sigma_k^i + \sum_{k=1}^K n_k^i (\mu_k^i - \mu_i)(\mu_k^i - \mu_i)^T \right).
\end{split}
\end{equation}
\end{small}
where each client's contribution to \(\Sigma_i\) is determined by the coefficient \( \frac{N_k^i}{N_i} \), By performing eigenvalue decomposition on \( \Sigma_i \), we can derive the geometric shape of the global data distribution for class \( i \) without sharing any data. For more details on privacy constraints, please refer to Appendix \ref{Privacy Constraints}.

\subsection{Global Geometry-Guided Embedding Uncertainty Representation (GGEUR)}
\label{sec3.2}

Given $K$ clients and a classification task with $C$ classes, we describe below how geometric knowledge can be used to simulate the ideal global data distribution locally in both single-domain and multi-domain scenarios.

\vspace{-5pt}
\subsubsection{Single-Domain Federated Learning}
\label{sec3.2.1}

When the data originates from a single domain, we primarily address the issue of label skew. Before federated learning begins, each client uses CLIP to extract \( p \)-dimensional embeddings of its local data. Each client then locally computes the per-class local covariance matrices and local sample means, which are subsequently uploaded to the server. For each class, the server generates a global covariance matrix, \( \Sigma_1, \Sigma_2, \ldots, \Sigma_C \), using Equation (\textcolor{red}{4}). Next, each class’s global geometric shape is quantified based on Equation (\textcolor{red}{2}), and these global geometric shapes (i.e., eigenvalues and eigenvectors) are sent to each client.

\begin{figure}[t]
  \centering
   \includegraphics[width=1\linewidth]{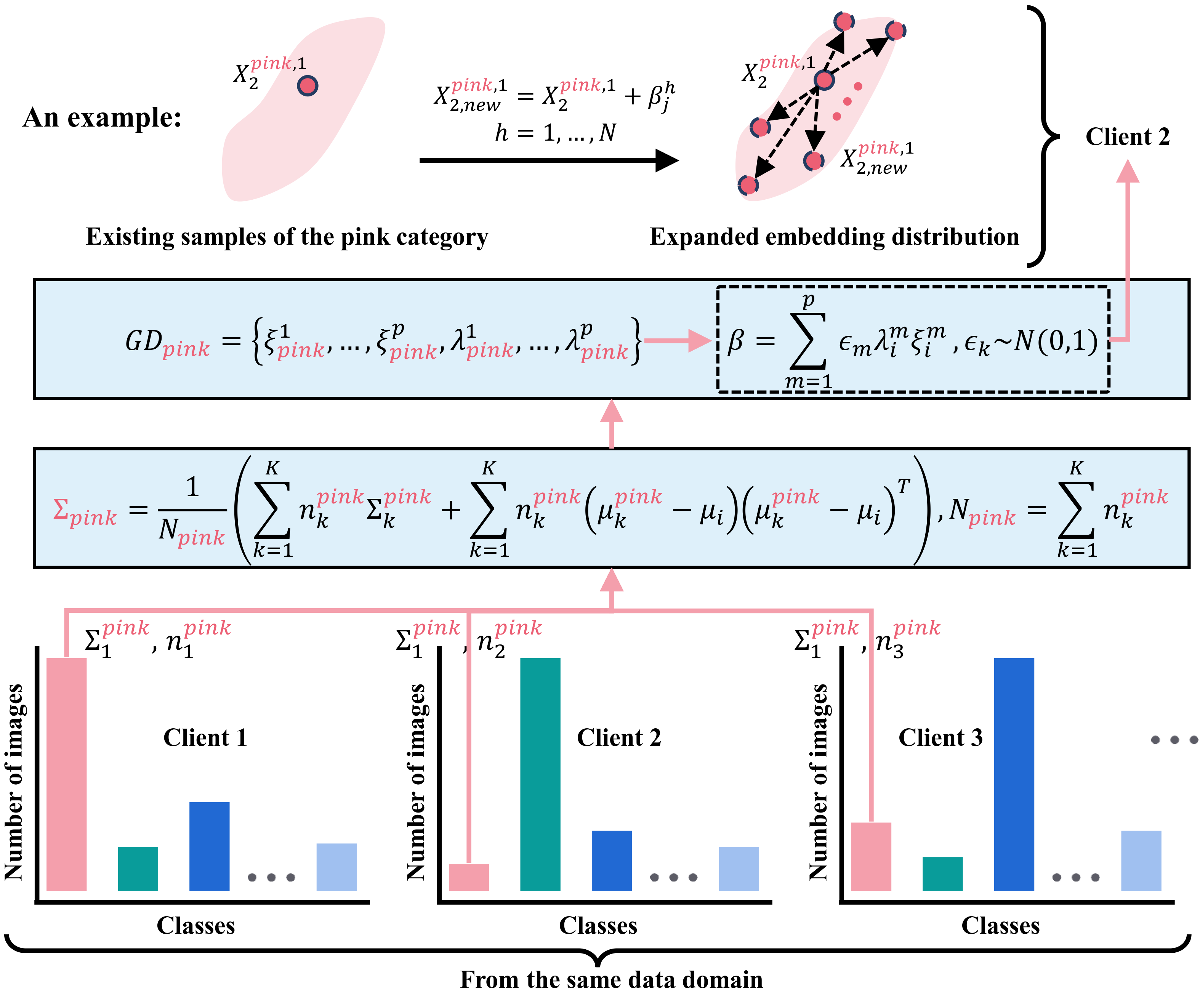}
\vskip -0.1in
   \caption{In a single-domain scenario, each class's global geometric shape is used to guide sample augmentation on each client. The example shows how new samples are generated for Client 2.}
   \label{fig3}
\vskip -0.1in
\end{figure}

We propose a Global Geometry-Guided Embedding Uncertainty Representation (\textbf{GGEUR)} for locally augmenting samples (i.e., embeddings) to simulate the global distribution. Specifically, for class \( i \), its global geometric shape can be represented as \( \text{GD}_i = \{\xi_i^1, \ldots, \xi_i^p, \lambda_i^1, \ldots, \lambda_i^p\} \). Suppose client \( k \) has only a limited number of class \( i \) embeddings, denoted as \( X_k^i = [X_k^{(i,1)}, \ldots, X_k^{(i,n_k^i)}] \in \mathbb{R}^{p \times n_k^i} \), where \( n_k^i \) represents the sample count. As shown in Figure \ref{fig3}, we first perform \( n_k^i \times N \) random linear combinations of the eigenvectors \( \xi_i^1, \ldots, \xi_i^p \) to produce \( n_k^i \times N \) different vectors $\beta = \sum_{m=1}^p \epsilon_m \lambda_i^m \xi_i^m \in \mathbb{R}^p$, where \( \epsilon_j \) follows a standard Gaussian distribution \( N(0, 1) \) and is scaled by the eigenvalues \( \lambda_i^j \) to control the magnitude. These vectors provide the direction and range for simulating the global distribution.
Next, for each existing sample \( X_k^{(i,1)}, \ldots, X_k^{(i,n_k^i)} \), we apply \( N \) of these vectors to generate \( n_k^i \times N \) new samples as follows:
\begin{equation}
\begin{split}
X_{(k,h)}^{(i,j)} = X_k^{(i,j)} + \beta_j^h,  j = 1, \ldots, n_k^i,  h = 1, \ldots, N.
\end{split}
\end{equation}
The total number of new samples can be set based on task requirements; in this study, for each local class, we ensure the total number of new and existing samples reaches $2000$. Figure \ref{fig3} and Algorithm \ref{alg1} illustrate this process in detail. After embedding space augmentation, each client trains an MLP as the local model.

\begin{algorithm}[t]
\caption{GGEUR (Single-Domain Scenario)}
\label{alg1}
\begin{algorithmic}[1]
\Require $X_k^i = [X_k^{(i,1)}, \dots, X_k^{(i, n_k^i)}] \in \mathbb{R}^{p \times n_k^i}$: Sample set of class $i$ at client $k$, 
$GD_i = \{\xi_i^1, \dots, \xi_i^p, \lambda_i^1, \dots, \lambda_i^p\}$: Global geometric shape (eigenvectors and eigenvalues) of class $i$, 
$N$: Number of new samples to generate per original sample
\Ensure $X_{\text{new}}^i$: Augmented sample set of class $i$ at client $k$

\Function{GGEUR}{$X, GD_i, N$}
    \State $X_{\text{gen}} \gets \emptyset$ \textcolor{cvprblue}{\Comment{Initialize generated samples}}
    \For {$h = 1$ to $N$}
        \State $\beta^h \gets \sum_{m=1}^{p} \epsilon_m \lambda_i^m \xi_i^m, \epsilon_m \sim \mathcal{N}(0,1)$ 
        \State \textcolor{cvprblue}{\Comment{Generate new vector}}
        \State $X_{\text{gen}} \gets X_{\text{gen}} \cup \{X + \beta^h\}$ 
      \State \textcolor{cvprblue}{\Comment{Generate and add new sample, Equation (5)}}
    \EndFor
    \State \textbf{return} $X_{\text{gen}}$
\EndFunction

\State $X_{\text{new}}^i \gets \emptyset$ \textcolor{cvprblue}{\Comment{Initialize augmented sample set}}
\For {$j = 1$ to $n_k^i$}
    \State $X_{\text{new}}^i \gets X_{\text{new}}^i \cup \textsc{GGEUR}(X_k^{(i,j)}, GD_i, N)$
\EndFor
\State \textbf{return} $X_{\text{new}}^i$
\end{algorithmic}
\end{algorithm}

\vspace{-5pt}
\subsubsection{Multi-Domain Federated Learning}
\label{sec3.2.2}

\begin{figure}[h]
\vskip -0.05in
  \centering
   \includegraphics[width=1\linewidth]{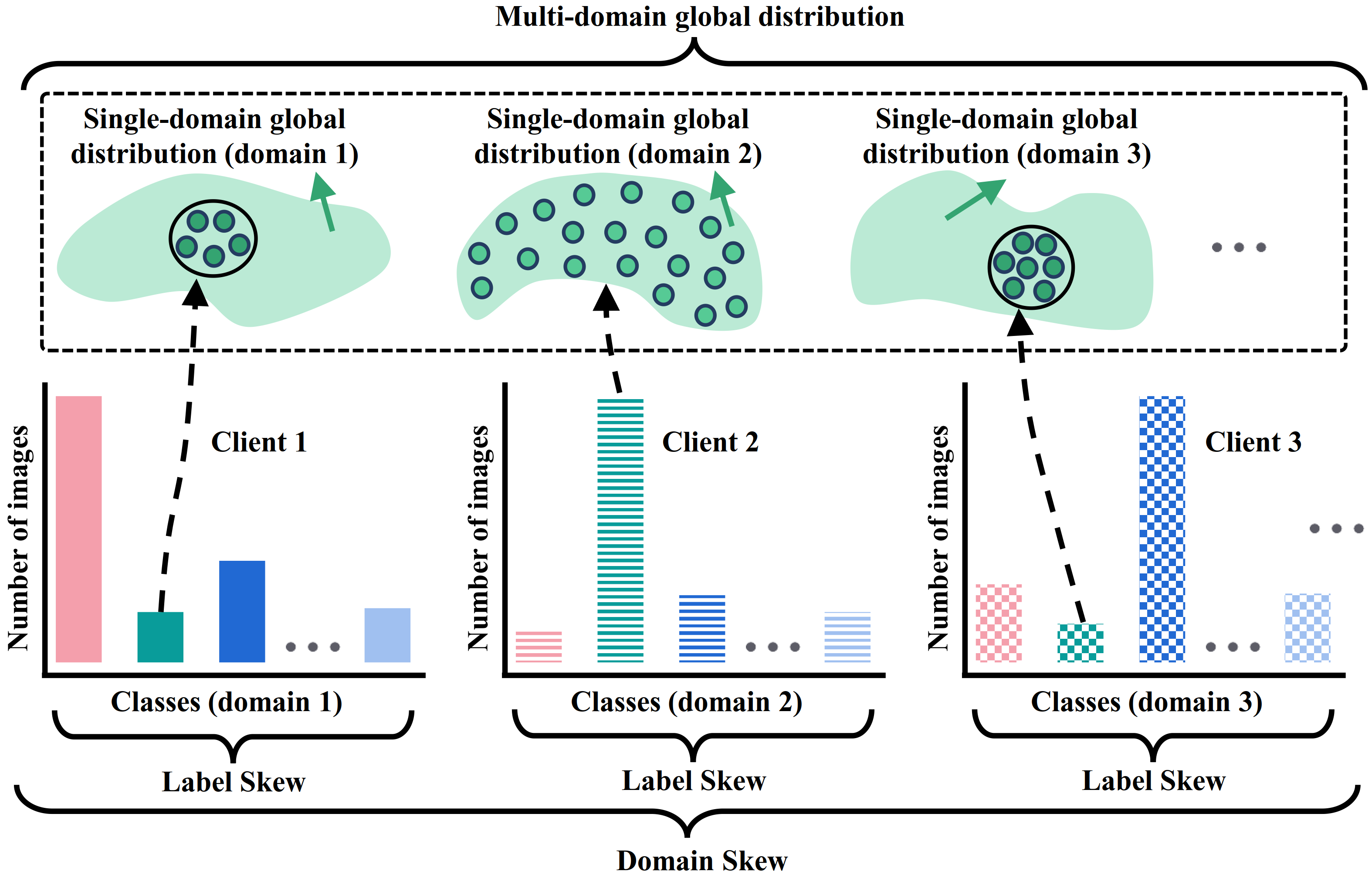}
\vskip -0.05in
   \caption{Federated scenario with both label skew and domain skew. Different textures represent data from distinct domains.}
   \label{fig4}
\vskip -0.1in
\end{figure}

When data originates from multiple domains, we encounter a complex scenario where both label skew and domain skew exist, as illustrated in Figure \ref{fig4}. In this case, different clients possess data from different domains. For clarity, we introduce two new concepts: the single-domain global distribution and the multi-domain global distribution.

\begin{definition}[Single-Domain Global Distribution]
Within a particular data domain, when a category’s sample count is sufficient and diverse enough to fully represent the category, this distribution of samples is termed the \textbf{single-domain global distribution} for that category. For example, in Figure \ref{fig4}, client 2’s \textcolor{forestgreen}{green samples} are numerous and adequately cover the global distribution of the \textcolor{forestgreen}{green category} within domain 2.
\end{definition}

\begin{definition}[Multi-Domain Global Distribution]
For a given category, the combined single-domain global distributions from all data domains constitute the \textbf{multi-domain global distribution} of that category. As shown in Figure \ref{fig4}, the \textcolor{forestgreen}{green distributions} across all domains together form the multi-domain global distribution for the \textcolor{forestgreen}{green category}.
\end{definition}

Clearly, addressing both label skew and domain skew requires two steps: \textbf{(1)} Local sample augmentation to simulate the single-domain global distribution within each client’s data domain. For example, in Figure \ref{fig4}, client 1 lacks sufficient green samples to represent the single-domain global distribution and thus requires sample augmentation. \textbf{(2)} Generation of new samples on each client to simulate the single-domain global distributions of other data domains.

\begin{figure}[h]
\vskip -0.05in
  \centering
   \includegraphics[width=1\linewidth]{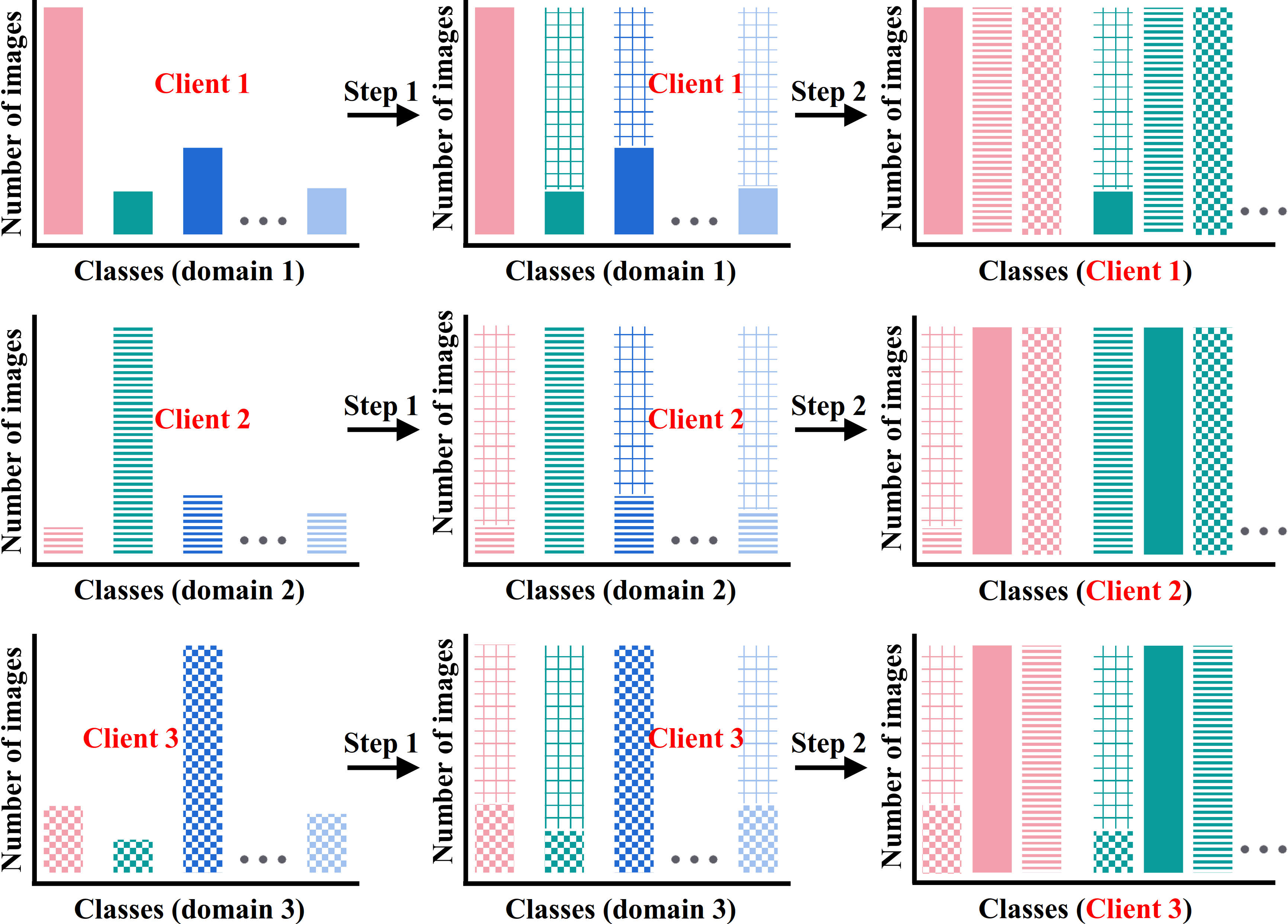}
\vskip -0.05in
   \caption{Example with 3 clients: Step 1 generates samples for each client from its own domain, while Step 2 simulates samples from other domains for each client.}
   \label{fig5}
\end{figure}

\begin{figure*}[t]
  \centering
   \includegraphics[width=1\linewidth]{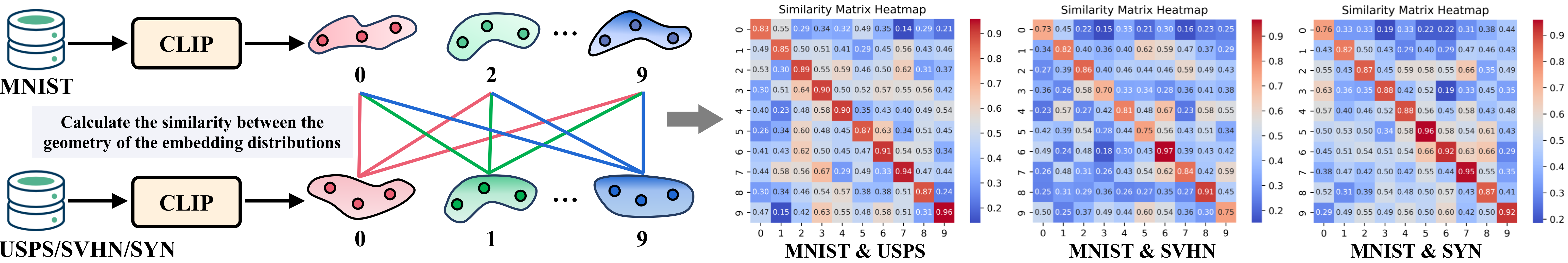}
\vskip -0.05in
   \caption{Study of geometric shape similarity on the Digits dataset.}
   \label{fig6}
\vskip -0.1in
\end{figure*}

As depicted in Figure \ref{fig5}, Step 1 addresses the label skew on each client, while Step 2 addresses domain skew. In summary, our objective is to simulate the ideal multi-domain global distribution for each category locally on each client. Unlike single-domain federated learning, completing Step 1 in this context poses the challenge of obtaining the geometric shape of the single-domain global distribution. For instance, in Figure \ref{fig5}, client 1 has only a small number of green samples, making it difficult to access the single-domain global distribution for the green category. Fortunately, \textbf{we found} that when embeddings are extracted using CLIP, the geometric shapes of multiple single-domain global distributions corresponding to the same category across different domains are similar. This finding implies that we can identify a cross-domain shared geometric shape for each category, representing its global geometric shape.

Specifically, we investigate this using the Digits dataset, which includes four digit-recognition datasets (i.e., MNIST \cite{MNIST_IEEE98}, USPS \cite{USPS_PAMI94}, SVHN \cite{svhn_NeurIPS11}, and SYN \cite{syn}), each representing a different domain. First, we use CLIP (ViT-B/16) \cite{clip} to extract image embeddings for all categories across the four datasets. Then, we compute the similarity between the geometric shapes of the embedding distributions for each category in MNIST and the corresponding categories in USPS, SVHN, and SYN, as shown in Figure \ref{fig6}. The significantly higher values in the diagonal of each heatmap indicate that, across domains, the geometric shapes of embeddings for the same category exhibit a consistent pattern, rather than being randomly distributed. This allows us to simplify the multi-domain label skew problem in Step 1 to a label skew problem in the single-domain scenario.

In detail, we first extract embeddings using CLIP and then locally calculate the covariance matrix and mean of local samples, which are uploaded to the server. Given CLIP's cross-domain consistency in representing the same class, we continue to use Equation (\textcolor{red}{4}) on the server to produce a shared global covariance matrix $\Sigma_1, \Sigma_2, \dots, \Sigma_C$ for each class. We further obtain the shared global geometric shapes, which are then sent to clients. Each client then applies GGEUR to augment samples, thereby simulating the single-domain global distribution locally. 
\textbf{Following this}, clients also need to generate new samples to simulate the single-domain global distribution from other data domains. Although the geometric shapes of embedding distributions across different domains are similar, the positions of these distributions differ. We propose transferring the class prototypes (local sample means) from other domains to the local client, and applying GGEUR to these prototypes to generate new samples, thus simulating the multi-domain global distribution. Algorithm \ref{alg2} details this process. For each client, in Step 1, we ensure that the total number of new and existing samples for each class is $500$. Step 2 generates $500$ new samples based on each prototype separately.

\subsection{Comparison with Analogous Methods}

Federated data augmentation aim to bridge the gap between local data distributions and the ideal global distribution by generating more diverse data samples on clients \cite{hao2021towards,shin2020xor,ma2023feature}. For example, FedMix \cite{yoon2021fedmix} and FEDGEN \cite{zhu2021data} use MixUp and its variants to augment client data, thus mitigating label skew. However, due to the lack of knowledge-based guidance, these methods largely rely on the diversity of local data. FedFA \cite{zhou2023fedfa} assumes local data follows a Gaussian distribution and generates new samples centered on class prototypes. Nevertheless, the Gaussian assumption is overly idealistic \cite{ma2024geometric}, limiting the ability of generated samples to adequately reduce the discrepancy between local and global distributions. \textbf{Compared to} the Gaussian assumption, the geometric shapes proposed in this work provide a more accurate description of embedding distributions. GGEUR estimates the geometric shape of the global distribution without compromising privacy and leverages it to guide data augmentation on clients. The introduction of additional geometric knowledge enables our method to effectively reduce the gap between local and global distributions.

\section{Experiments}
\label{sec4}

In this section, we conduct a comprehensive evaluation of our method under scenarios with label skew, domain skew, and the coexistence of both.

\subsection{Experiment Setup}
\label{sec4.1}

\textbf{Label Skew Datasets.} We evaluate our method on three single-domain image classification tasks.
\begin{itemize}[]
    \item Cifar-10 \cite{cifar} contains $10$ classes, with $50,000$ images for training and $10,000$ images for validation.
    \item Cifar-100 \cite{cifar} covers $100$ classes, with $50,000$ training images and $10,000$ validation images.
\item Tiny-ImageNet \cite{tiny} is the subset of ImageNet with $100K$ images of size $64 \times 64$ with $200$ classes scale.
\end{itemize}

\noindent \textbf{Domain Skew Datasets.} We evaluated our method on the multi-domain image classification dataset Digits and conducted analyses on Office-Caltech and PACS.
\begin{itemize}[]
    \item Digits \cite{MNIST_IEEE98,USPS_PAMI94,svhn_NeurIPS11,syn} includes four domains: MNIST, USPS, SVHN and SYN, each with $10$ categories.
    \item Office-Caltech \cite{office_caltech} includes four domains: Caltech, Amazon, Webcam, and DSLR, each with $10$ categories.
\item PACS \cite{pacs} includes four domains: Photo (P) with $1, 670$ images, Art Painting (AP) with $2,048$ images, Cartoon (Ct) with $2,344$ images and Sketch (Sk) with $3,929$ images. Each domain holds seven categories.
\end{itemize}

\begin{figure}[h]
\vskip -0.15in
\centering
	\begin{minipage}{0.495\linewidth}
		\centering
		\includegraphics[width=1\linewidth]{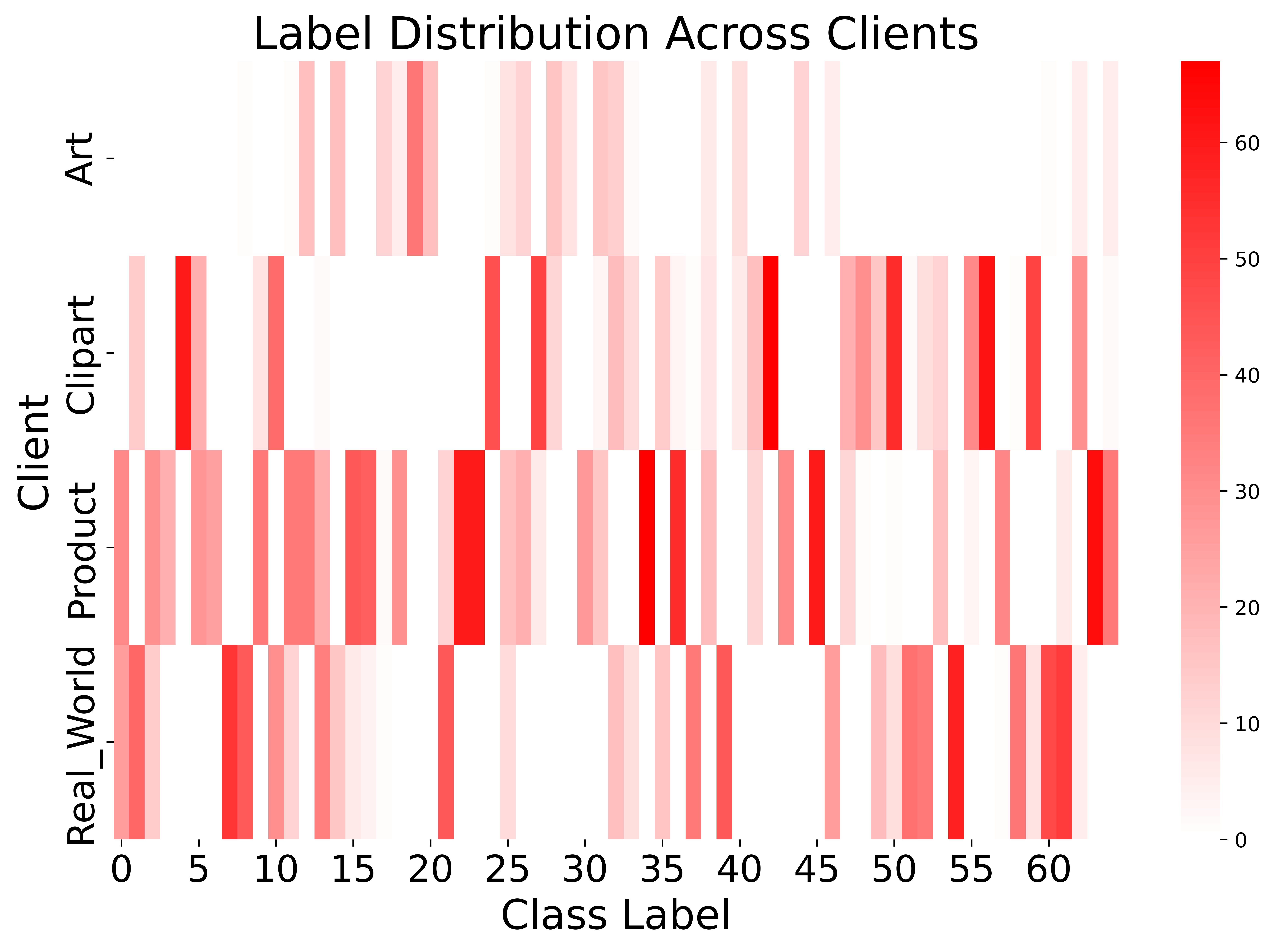}
	\end{minipage}
	\begin{minipage}{0.495\linewidth}
		\centering
		\includegraphics[width=1\linewidth]{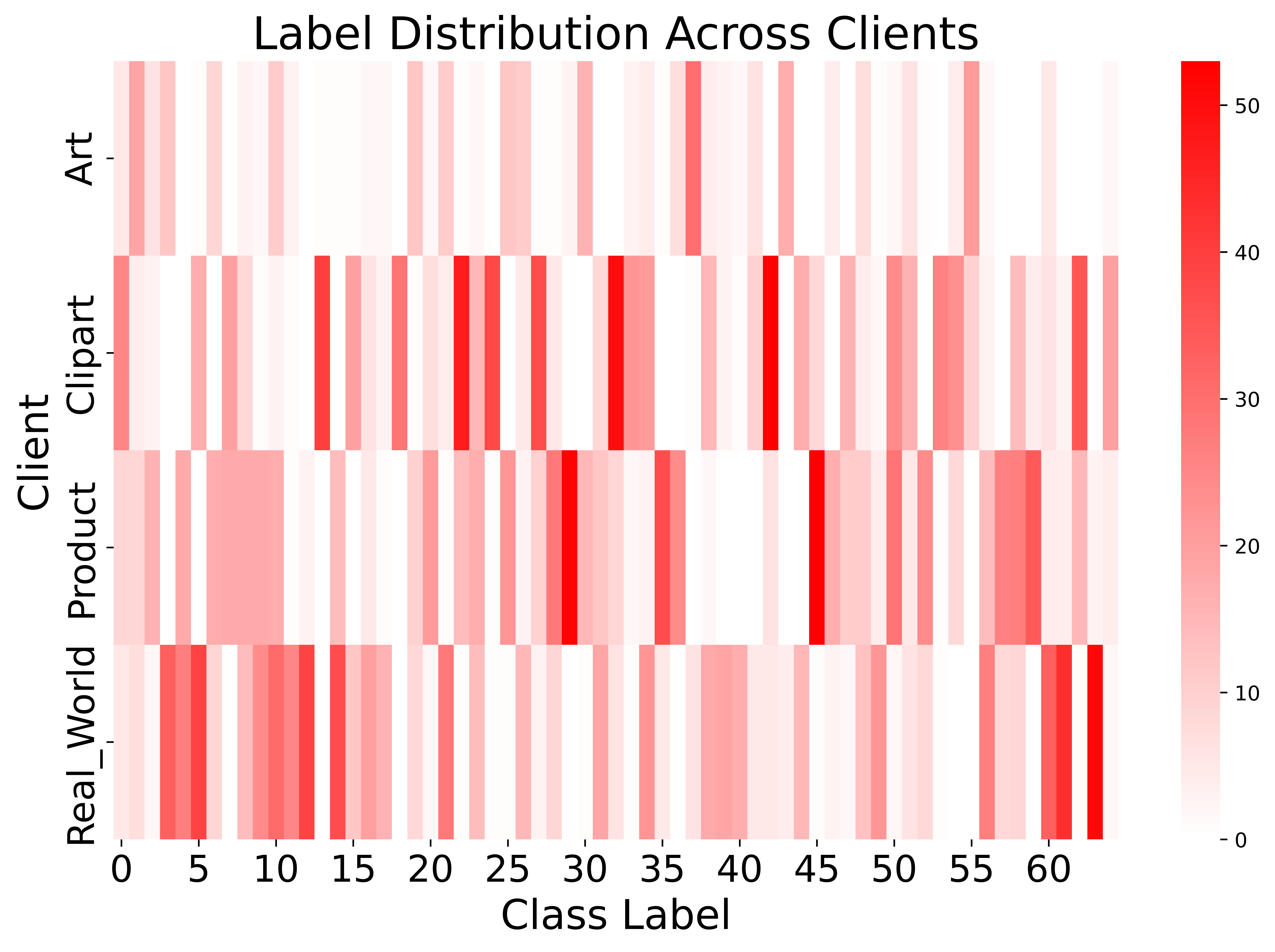}
	\end{minipage}
\vskip -0.1in
\caption{Number of samples per class across four clients when $\beta$ equals $0.1$ and $0.5$, with \textbf{each client holding data from a different domain.} Additional cases are provided in Appendix \ref{Dataset}.}
\label{fig7}
\vskip -0.15in
\end{figure}

\noindent \textbf{Dataset with Coexisting Label Skew and Domain Skew.}
Office-Home \cite{office_home} includes $4$ domains: Art (A), Clipart (C), Product (P), and Real World (R), each containing $65$ classes. To increase the challenge, we designed a new partitioning method for the multi-domain dataset Office-Home to create a scenario where label skew and domain skew coexist. For details, please refer to Appendix \ref{Dataset}.
We name the newly constructed dataset \textbf{Office-Home-LDS (Label and Domain Skew)}. Figure \ref{fig7} shows the data distribution of Office-Home-LDS with $\beta=0.1$ and $\beta=0.5$.

\noindent \textbf{Implementation Details and Comparison Methods.} On the label skew dataset, we applied GGEUR to FedAvg \cite{fedavg} using CLIP (ViT-B/16) \cite{clip} as the backbone network and compared it with state-of-the-art federated data augmentation methods in heterogeneous federated learning, including FedMix \cite{yoon2021fedmix}, FEDGEN \cite{zhu2021data}, and FedFA \cite{zhou2023fedfa}. Additionally, we compared it with other advanced methods that use CLIP (ViT-B/16) as the backbone, such as FedTPG \cite{fedtpg} and FedCLIP \cite{fedclip}. On the domain skew and Office-Home-LDS datasets, we focused on exploring the enhancement effect of GGEUR across various federated architectures. Therefore, we applied GGEUR to FedAvg, SCAFFOLD \cite{SCAFFOLD}, MOON \cite{moon}, FedDyn \cite{acar2021federated}, FedOPT \cite{fedopt}, FedProto \cite{fedproto}, and FedNTD \cite{lee2022preservation}, all of which use CLIP for image feature extraction and \textbf{a single-layer MLP} as the local model. All experimental settings for FL methods are consistent with the latest benchmark \cite{fl3}; details are provided in Appendix \ref{Implementation Details}.

\noindent \textbf{Evaluation Metrics.} 
Following \cite{li2019fair}, we utilize the Top-1 (\%) accuracy and the standard deviation of accuracy across multi-domains as evaluation metrics. A smaller standard deviation indicates better Performance Fairness across different domains. We use the average results from the last five rounds accuracy and variance as the final performance.

\begin{table}[t]
\small
\setlength{\abovecaptionskip}{0cm}
\centering
\setlength{\tabcolsep}{2.4pt} 
\renewcommand\arraystretch{0.98}
\caption{Comparison results on CIFAR-100 and Tiny-ImageNet datasets with different degrees of label skew ($\beta$ values). The best results are shown in \underline{\textbf{underlined bold}}. FedAvg (CLIP+MLP) indicates that the backbone network uses CLIP and MLP, and the federated learning method employs FedAvg.}
\scalebox{0.96}{
\begin{tabular}{r||ccc|ccc}
\hline\thickhline
\rowcolor{lightgray}
& \multicolumn{3}{c}{CIFAR-100} & \multicolumn{3}{c}{Tiny-ImageNet} \\
\cline{2-7}
\rowcolor{lightgray}
\multirow{-2}{*}{Methods} 
 & 0.5 & 0.3 & 0.1  & 0.5 & 0.3 & 0.1  \\
\hline\hline

Zero-Shot CLIP & \multicolumn{3}{c|}{64.87}    & \multicolumn{3}{c}{63.67}   \\
FedTPG   & 71.40 & 70.95 & 68.63   & 67.63 & 66.72 & 64.71  \\
FedCLIP   & 72.03 & 71.20 & 70.64   & 70.41 & 70.37 & 69.50  \\
\hline \hline
FedMix (CLIP+MLP) & 81.31 & 79.62 & 73.85   & 70.89 & 68.57 & 63.43  \\
FEDGEN (CLIP+MLP) & 81.24 & 78.97 & 73.15   & 72.37 & 70.35 & 64.16  \\
FedFA (CLIP+MLP) & 81.98 & 79.31 & 74.68   & 70.41 & 70.68 & 64.62  \\ \hline 
FedAvg (CLIP+MLP)  & 81.41 & 77.68 & 68.22  & 70.08 &67.65  & 60.10  \\
\textbf{+ GGEUR}  & \underline{\textbf{83.31}} & \underline{\textbf{81.65}} & \underline{\textbf{77.70}}   & \underline{\textbf{73.89}} & \underline{\textbf{72.19}} & \underline{\textbf{66.86}}  \\
\bottomrule \hline
\end{tabular}}
\label{tab1}
\vskip -0.15in
\end{table}

\begin{table}[t]
\small
\setlength{\abovecaptionskip}{0cm}
\centering
\setlength{\tabcolsep}{4.1pt} 
\renewcommand\arraystretch{0.98}
\caption{Evaluation results of GGEUR on CIFAR-10 and CIFAR-100 with more severe label skew (i.e., smaller $\beta$ values).}
\scalebox{1}{
\begin{tabular}{r||ccccc}
\hline\thickhline
\rowcolor{lightgray}
& \multicolumn{5}{c}{CIFAR-10}  \\
\cline{2-6}
\rowcolor{lightgray}
\multirow{-2}{*}{Methods} 
 & 0.01 & 0.03 & 0.05  & 0.07 & 0.09   \\
\hline\hline
FedAvg (CLIP+MLP) & 90.87 & 90.13 & 91.96   &92.05 & 91.82 \\
\textbf{+ GGEUR}  & \underline{\textbf{94.39}} & \underline{\textbf{94.25}} & \underline{\textbf{95.07}}   & \underline{\textbf{95.21}} & \underline{\textbf{95.38}}  \\
\bottomrule 

\rowcolor{lightgray}
& \multicolumn{5}{c}{CIFAR-100}  \\
\cline{2-6}
\rowcolor{lightgray}
\multirow{-2}{*}{Methods} 
 & 0.01 & 0.03 & 0.05  & 0.07 & 0.09   \\
\hline\hline
FedAvg (CLIP+MLP) & 58.71 & 60.77 & 62.32   & 61.69 & 66.51 \\
\textbf{+ GGEUR}  & \underline{\textbf{75.72}} & \underline{\textbf{75.40}} & \underline{\textbf{75.96}}   & \underline{\textbf{76.72}} & \underline{\textbf{78.00}}  \\
\bottomrule \hline
\end{tabular}}
\label{tab2}
\vskip -0.15in
\end{table}

\begin{table}[t]
\small
\setlength{\abovecaptionskip}{0cm}
\centering
\setlength{\tabcolsep}{4pt} 
\renewcommand\arraystretch{0.98}
\caption{Evaluation results of GGEUR on Tiny-ImageNet with more severe label skew (i.e., smaller $\beta$ values).}
\scalebox{1}{
\begin{tabular}{r||ccccc}
\hline\thickhline
\rowcolor{lightgray}
& \multicolumn{5}{c}{Tiny-ImageNet}  \\
\cline{2-6}
\rowcolor{lightgray}
\multirow{-2}{*}{Methods} 
 &0.01  &0.03  & 0.05 &0.07 & 0.09     \\
\hline\hline
FedAvg (CLIP+MLP) & 53.03 & 54.57  & 58.91  &58.77 &59.13     \\
\textbf{+ GGEUR} & \underline{\textbf{64.27}}  & \underline{\textbf{65.79}}   & \underline{\textbf{66.49}} & \underline{\textbf{66.34}}  & \underline{\textbf{66.85}}   \\
\bottomrule \hline
\end{tabular}}
\label{tab3}
\vskip -0.1in
\end{table}

\subsection{Evaluation Results on Label Skew Dataset}
\label{sec4.2}

\noindent \textbf{Main Results.} Tables \ref{tab1}, \ref{tab2}, and \ref{tab3} show the performance improvement of GGEUR over FedAvg (CLIP+MLP) \cite{fedavg} under different $\beta$ values, along with comparison results with other methods. Under all label distribution skew settings, GGEUR significantly outperforms other methods, demonstrating superior classification accuracy and adaptability to imbalanced label distributions. Particularly at lower $\beta$ values (i.e., with more severe label skew), GGEUR notably enhances the performance of FedAvg (CLIP+MLP) on CIFAR-100, validating its effectiveness in handling extreme label skew scenarios. For example, when $\beta$ is $0.01$, $0.03$, and $0.05$, GGEUR achieves performance gains of $\textbf{17.01\%}$, $\textbf{14.63\%}$, and $\textbf{13.64\%}$, respectively. These results indicate that GGEUR not only exhibits good generalization ability in standard settings but also maintains robust performance in cases of significant label skew. Moreover, our large-scale client experiments further confirm the scalability and consistent effectiveness of GGEUR under increased client participation, details are provided in Appendix \ref{Large-Scale Client}.

\noindent \textbf{Comparison with Peer Methods.} We implemented FedMix, FEDGEN, and FedFA for comparison in Table \ref{tab1}, and in all cases, our method outperformed these three approaches. This demonstrates the superiority of geometric shapes as knowledge over the Gaussian assumption and traditional data augmentation methods.

\begin{table}[t]
\small
\setlength{\abovecaptionskip}{0cm}
\centering
\setlength{\tabcolsep}{0.3pt} 
\renewcommand\arraystretch{0.95}
\caption{Evaluation results on the Digits dataset. GGEUR by default includes both Step 1 and Step 2.}
\scalebox{0.96}{
\begin{tabular}{r||cccc|cc}
\hline\thickhline
\rowcolor{lightgray}
& \multicolumn{6}{c}{Digits}  \\
\cline{2-7}
\rowcolor{lightgray}
\multirow{-2}{*}{Methods} 
 &MNIST  &USPS  &SVHN &SYN & AVG $\uparrow$ &STD $\downarrow$    \\
\hline\hline
FedMix (CLIP+MLP)  &95.03  &90.25 & 57.50 &72.60  &78.85  &14.89  \\
FEDGEN (CLIP+MLP) &95.85  &92.52  &58.77 &73.62  &80.19  &14.99  \\
FedFA (CLIP+MLP) &96.68  &92.97  &57.87  &75.53  &80.76  &15.44  \\ \hline  \hline 

FedAvg \cite{fedavg}  &90.40  &60.30  & 34.68 &46.99  & 58.09 & 20.74     \\
FedAvg (CLIP+MLP)  &95.12  &89.74  & 56.36 &65.17  & 76.60 &16.25   \\
\rowcolor{cvpryellow!10}
+ \textbf{GGEUR (Step 1)}  &96.02  &93.02  & 58.55 &73.13  & 80.18 &15.28   \\
\rowcolor{cvpryellow!25}
+ \textbf{GGEUR (Step 1 \& 2)}  &97.05  &94.12  &63.54 &74.73  & \underline{\textbf{82.36}} &\underline{\textbf{13.84}}   \\ \hline

SCAFFOLD \cite{SCAFFOLD} & 97.79 &94.45 & 26.64 & 90.69 & 77.39 &29.41   \\
SCAFFOLD (CLIP+MLP)  & 94.62 & 90.08 & 54.33 & 68.71 & 76.93 &16.31   \\
+ \textbf{GGEUR}  & 95.91 & 92.08 & 63.25 & 71.54 & \underline{\textbf{80.70}}  &\underline{\textbf{13.69}}  \\ \hline

MOON \cite{moon} & 92.78 & 68.11 & 33.36 & 39.28 & 58.36  &23.82   \\
MOON (CLIP+MLP)  & 75.64 & 73.09 & 38.83 & 52.74 & 60.07  &\underline{\textbf{15.14}}   \\
\rowcolor{cvpryellow!10}
\textbf{+ GGEUR (Step 1)}  &84.64  &81.96  &43.04  &60.35  & 67.50 &16.97   \\
\rowcolor{cvpryellow!25}
\textbf{+ GGEUR (Step 1 \& 2)} & 95.16 & 91.13 & 55.23 & 71.00 & \underline{\textbf{78.13}}  &16.08   \\ \hline

FedDyn \cite{acar2021federated} & 88.91 & 60.34 & 34.57 & 50.72 & 58.65   &19.76   \\
FedDyn (CLIP+MLP) & 95.46 & 92.13 & 58.89 & 70.30 & 79.19  &15.19  \\
\textbf{+ GGEUR}   & 97.07 & 94.02 & 63.34 & 74.83 & \underline{\textbf{82.31}}  &\underline{\textbf{13.88}} \\ \hline

FedOPT \cite{fedopt} & 92.71 & 87.62 & 31.32 & 87.92 & 74.89  &25.38   \\
FedOPT (CLIP+MLP)  & 94.57 & 88.79 & 58.65 & 66.47 & 77.12 &14.96  \\
\textbf{+ GGEUR}   & 96.43 & 93.47 & 62.35 & 70.75 & \underline{\textbf{80.75}}   &\underline{\textbf{14.54}}  \\ \hline

FedProto \cite{fedproto} & 90.54 & 89.54 & 34.61 & 58.00 & 68.18  &23.38   \\
FedProto (CLIP+MLP)  & 94.86 & 92.63 & 54.29 & 65.52 & 76.83  &17.40   \\
\textbf{+ GGEUR}   & 97.19 & 94.12 & 63.70 & 73.83 & \underline{\textbf{82.21}} &\underline{\textbf{13.96}}  \\ \hline

FedNTD \cite{lee2022preservation} & 52.31 & 58.07 & 18.03 & 97.29 & 56.43  &28.12  \\
FedNTD (CLIP+MLP)  & 95.82 & 91.43 & 58.26 & 69.95 & 78.86   &15.41  \\
\textbf{+ GGEUR}  & 97.08 & 94.32 & 63.57 & 73.53 & \underline{\textbf{82.13}} &\underline{\textbf{14.06}}   \\ 

\bottomrule \hline
\end{tabular}}
\label{tab4}
\vskip -0.1in
\end{table}

\subsection{Evaluation Results on Domain Skew Dataset}
\label{sec4.3}

\noindent \textbf{Ablation Study.} As shown in Algorithm \ref{alg2}, simulating the multi-domain global distribution requires two steps. We selected the classic FedAvg (CLIP+MLP) and MOON (CLIP+MLP) for an ablation study on the Digits dataset. The experimental results, \textcolor{cvpryellow}{highlighted in yellow} in Table \ref{tab4}, show that each step incrementally improves the original methods, and their combination achieves the best results.

\noindent \textbf{Main Results.} Tables \ref{tab4}, \ref{tab5}, and \ref{tab6} present the experimental results on datasets with domain skews. It can be observed that simply using CLIP \cite{clip} for image representation, combined with a single-layer MLP for federated learning, already surpasses existing methods. This improvement is attributed to the advancements in the foundation model, and GGEUR can further significantly enhance the performance of the global model. For instance, on the Digits dataset, GGEUR improves the average performance of FedAvg (CLIP+MLP) \cite{fedavg}, MOON (CLIP+MLP) \cite{moon}, and FedProto (CLIP+MLP) \cite{fedproto} by $\textbf{5.76\%}$, $\textbf{18.06\%}$, and $\textbf{5.38\%}$, respectively. Additionally, compared to other methods, GGEUR significantly reduces the accuracy variance across different domains. This demonstrates that GGEUR effectively adapts to features from different domains when handling cross-domain data distribution disparities, providing more robust and fair performance. We also conducted experiments to measure the computational cost of GGEUR, which demonstrated that the additional overhead introduced by our method is minimal. Further details can be found in Appendix \ref{Computational Cost}.

On the Office-Caltech and PACS datasets, since the use of CLIP+MLP as the backbone network already achieves very high performance across all methods, these datasets are less challenging. However, our evaluation results can serve as a reference for other research.

\noindent \textbf{Comparison with Peer Methods.} We implemented FedMix (CLIP+MLP), FEDGEN (CLIP+MLP), and FedFA (CLIP+MLP) for comparison in Table \ref{tab4}. When GGEUR is applied to FedAvg (CLIP+MLP), it outperforms these three methods by $\textbf{3.51\%}$, $\textbf{2.17\%}$, and $\textbf{1.60\%}$, respectively.

\begin{table}[t]
\small
\setlength{\abovecaptionskip}{0cm}
\centering
\setlength{\tabcolsep}{1.15pt} 
\renewcommand\arraystretch{0.9}
\caption{Evaluation results on Office-Caltech.}
\scalebox{0.96}{
\begin{tabular}{r||cccc|cc}
\hline\thickhline
\rowcolor{lightgray}
& \multicolumn{6}{c}{Office-Caltech}  \\
\cline{2-7}
\rowcolor{lightgray}
\multirow{-2}{*}{Methods} 
 &Am  &Ca  &D &W & AVG $\uparrow$ &STD $\downarrow$    \\
\hline\hline
FedAvg \cite{fedavg}  & 81.99 & 73.21 & 79.37 & 67.93 & 75.62 & 6.31     \\
FedAvg (CLIP+MLP)  & 98.26 & 96.74 & 100 & 100 & \underline{\textbf{98.75}} &\underline{\textbf{1.57}}   \\ \hline

SCAFFOLD \cite{SCAFFOLD} & 39.77 & 42.50 & 78.02 & 70.69 & 57.75 &19.44   \\
SCAFFOLD (CLIP+MLP)  & 96.18 & 92.88 & 95.83 & 93.26 & \underline{\textbf{94.54}} &\underline{\textbf{1.70}}   \\ \hline

MOON \cite{moon} & 84.42 & 75.98 & 84.67 & 68.97 & 78.51  &7.53   \\
MOON (CLIP+MLP)  & 98.61 & 97.33 & 100 &98.88 & \underline{\textbf{98.70}}  &\underline{\textbf{1.09}}   \\ \hline

FedDyn \cite{acar2021federated} & 84.02 & 72.59 & 77.34 & 68.97 & 75.72   &6.50   \\
FedDyn (CLIP+MLP) & 98.61 & 96.74 & 100 & 100 & \underline{\textbf{98.84}}  &\underline{\textbf{1.54}}  \\  \hline

FedOPT \cite{fedopt}  & 79.05 & 71.96 & 89.34 & 74.48 & 78.71  &7.67   \\
FedOPT (CLIP+MLP)  & 98.26 & 97.33 & 100 & 100 &\underline{\textbf{ 98.90}} &\underline{\textbf{1.33}}  \\  \hline

FedProto \cite{fedproto} & 87.79 & 75.98 & 90.00 & 79.31 & 83.27  &6.70   \\
FedProto (CLIP+MLP)  & 98.26 & 96.74 & 100 & 100 & \underline{\textbf{98.75}}  &\underline{\textbf{1.57}}   \\   \hline

FedNTD \cite{lee2022preservation} & 10.95 & 10.89 & 14.67 & 10.34 & 11.71  &1.99  \\
FedNTD (CLIP+MLP)  & 97.92 & 96.14 & 100 & 100 & \underline{\textbf{98.51}}  &\underline{\textbf{1.86}}  \\

\bottomrule \hline
\end{tabular}}
\label{tab5}
\vskip -0.1in
\end{table}

\begin{table}[t]
\small
\setlength{\abovecaptionskip}{0cm}
\centering
\setlength{\tabcolsep}{1.1pt} 
\renewcommand\arraystretch{0.9}
\caption{Evaluation results on PACS.}
\scalebox{0.96}{
\begin{tabular}{r||cccc|cc}
\hline\thickhline
\rowcolor{lightgray}
& \multicolumn{6}{c}{PACS}  \\
\cline{2-7}
\rowcolor{lightgray}
\multirow{-2}{*}{Methods} 
 &P  &AP  &Ct &Sk & AVG $\uparrow$ &STD $\downarrow$    \\
\hline\hline
FedAvg \cite{fedavg}  & 76.09 & 64.19 & 83.50 & 89.40 & 78.30 & 9.41     \\
FedAvg (CLIP+MLP)  & 99.40 & 98.37 & 99.01 & 93.64 &\underline{\textbf{97.60}}  &\underline{\textbf{2.32}}   \\ \hline

SCAFFOLD \cite{SCAFFOLD} & 61.95 & 45.44 & 58.87 & 54.64 & 55.25 &6.22   \\
SCAFFOLD (CLIP+MLP)  & 92.42 & 81.63 & 80.68 & 87.28 & \underline{\textbf{85.50}} &\underline{\textbf{4.72}}   \\ \hline

MOON \cite{moon} & 74.44 & 64.19 & 83.92 & 89.17 & 77.93  &9.53   \\
MOON (CLIP+MLP)  & 99.60 & 99.02 & 99.43 & 93.89 &\underline{\textbf{97.99}}  &\underline{\textbf{2.37}}   \\ \hline

FedDyn \cite{acar2021federated} & 78.17 & 63.29 & 82.27 & 89.93 & 78.66  &9.70   \\
FedDyn (CLIP+MLP) & 99.40 & 98.37 & 99.01 & 93.55 & \underline{\textbf{97.58}}   &\underline{\textbf{2.36}}  \\  \hline

FedOPT \cite{fedopt} & 78.66 & 67.66 & 82.41 & 83.68 & 78.12 &6.31   \\
FedOPT (CLIP+MLP)  & 99.40 & 98.37 & 99.01 & 93.64 & \underline{\textbf{97.60}}  &\underline{\textbf{2.32}}  \\  \hline

FedProto \cite{fedproto}  & 85.63 & 73.69 & 83.57 & 91.14 & 83.51  &6.31   \\
FedProto (CLIP+MLP)  & 99.40 & 98.21 & 99.01 & 94.23 & \underline{\textbf{97.71}}  &\underline{\textbf{2.06}}   \\   \hline

FedNTD \cite{lee2022preservation}  & 16.77 & 18.23 & 28.47 & 93.18 & 39.16  &31.51  \\
FedNTD (CLIP+MLP)  & 99.40 & 98.54 & 99.29 & 92.28 & \underline{\textbf{97.38}}  &\underline{\textbf{2.96}}  \\

\bottomrule \hline
\end{tabular}}
\label{tab6}
\vskip -0.1in
\end{table}

\begin{table}[t]
\small
\setlength{\abovecaptionskip}{0cm}
\centering
\setlength{\tabcolsep}{1pt} 
\renewcommand\arraystretch{0.96}
\caption{Evaluation results on Office-Home-LDS ($\beta=0.1$).}
\scalebox{0.91}{
\begin{tabular}{r||cccc|cc}
\hline\thickhline
\rowcolor{lightgray}
& \multicolumn{6}{c}{Office-Home-LDS}  \\
\cline{2-7}
\rowcolor{lightgray}
\multirow{-2}{*}{Methods} 
 &A  &C  &P &R & AVG $\uparrow$ &STD $\downarrow$    \\
\hline\hline
FedAvg (CLIP+MLP) \cite{fedavg}  &65.29 & 58.17 & 80.56 & 76.53 & 70.14 & 8.89     \\
\textbf{+ GGEUR}  &78.33 & 79.01 & 90.17 & 88.46 & \underline{\textbf{83.99}}   &\underline{\textbf{5.36}}   \\ \hline


SCAFFOLD (CLIP+MLP) \cite{SCAFFOLD} & 68.72 & 66.79 & 83.63 & 80.12 & 74.82 &7.20   \\
\textbf{+ GGEUR}  & 78.60 & 78.32 & 89.86 & 89.07 & \underline{\textbf{83.96}} &\underline{\textbf{5.51}}   \\ \hline

MOON (CLIP+MLP) \cite{moon} &69.27 & 68.63 & 86.56 & 82.87 & 76.83 &7.99   \\
\textbf{+ GGEUR}  & 72.02 & 70.31 & 86.11 & 83.87 & \underline{\textbf{78.08}}  &\underline{\textbf{6.98}}   \\ \hline

FedDyn (CLIP+MLP) \cite{acar2021federated} &58.30 & 55.19 & 77.63 & 72.86 & 65.99 &9.47   \\
\textbf{+ GGEUR} &78.88 & 78.55 & 90.47 & 88.46 & \underline{\textbf{84.09}}  &\underline{\textbf{5.42}}  \\  \hline

FedOPT (CLIP+MLP) \cite{fedopt} &58.44 & 54.89 & 76.80 & 72.25 & 65.59  &9.16   \\
\textbf{+ GGEUR}  & 79.01 & 78.32 & 90.84 & 88.61 & \underline{\textbf{84.20}}  &\underline{\textbf{5.59}}  \\  \hline

FedProto (CLIP+MLP) \cite{fedproto}  & 65.84 & 56.49 & 80.41 & 74.85 & 69.40   &9.09   \\
\textbf{+ GGEUR}  &78.05 & 77.71 & 89.79 & 87.84 & \underline{\textbf{83.35}}  &\underline{\textbf{5.51}}   \\   \hline

FedNTD (CLIP+MLP) \cite{lee2022preservation}  &69.68 & 66.64 & 84.53 & 80.96 & 75.46 &7.48  \\
\textbf{+ GGEUR}  &78.19 & 74.66 & 90.24 & 86.77 & \underline{\textbf{82.46}}  &\underline{\textbf{6.29}}  \\

\bottomrule \hline
\end{tabular}}
\label{tab7}
\vskip -0.15in
\end{table}

\subsection{Evaluation Results on Office-Home-LDS}
\label{sec4.4}

To fully demonstrate the potential of GGEUR, we constructed a more challenging dataset, Office-Home-LDS, which incorporates both label skew and domain skew. Office-Home-LDS simulates a more realistic and complex scenario of distributional imbalance, encompassing cross-domain data distribution differences and label imbalance. In Table \ref{tab7}, we show the remarkable performance gains of GGEUR over existing methods on this dataset, along with significant reductions in accuracy variance across different domains, highlighting its effectiveness in handling highly heterogeneous data scenarios. Specifically, GGEUR improved the average performance of FedAvg (CLIP+MLP) \cite{fedavg}, SCAFFOLD (CLIP+MLP) \cite{SCAFFOLD}, FedDyn (CLIP+MLP) \cite{acar2021federated}, FedOPT (CLIP+MLP) \cite{fedopt}, FedProto (CLIP+MLP) \cite{fedproto}, and FedNTD (CLIP+MLP) \cite{lee2022preservation} by $\textbf{13.85\%}$, $\textbf{9.14\%}$, $\textbf{18.1\%}$, $\textbf{18.61\%}$, $\textbf{13.95\%}$, and $\textbf{7.0\%}$, respectively. These results demonstrate that GGEUR can achieve robust performance on complex multi-domain, multi-class datasets, making it an effective approach for addressing both label and domain skew.

\begin{figure}[h]
\vskip -0.1in
\centering
	\begin{minipage}{0.495\linewidth}
		\centering
		\includegraphics[width=1\linewidth]{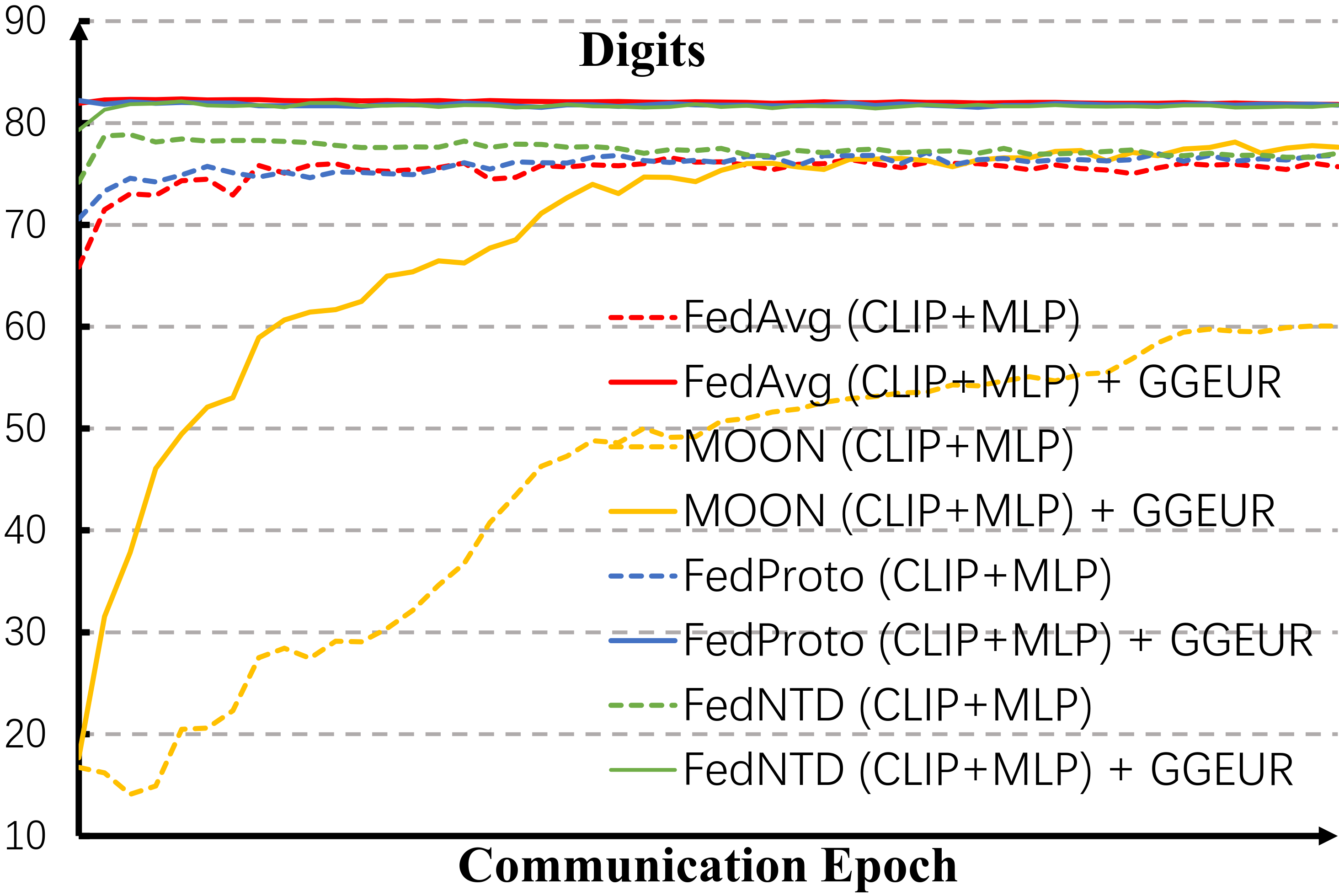}
	\end{minipage}
	\begin{minipage}{0.495\linewidth}
		\centering
		\includegraphics[width=1\linewidth]{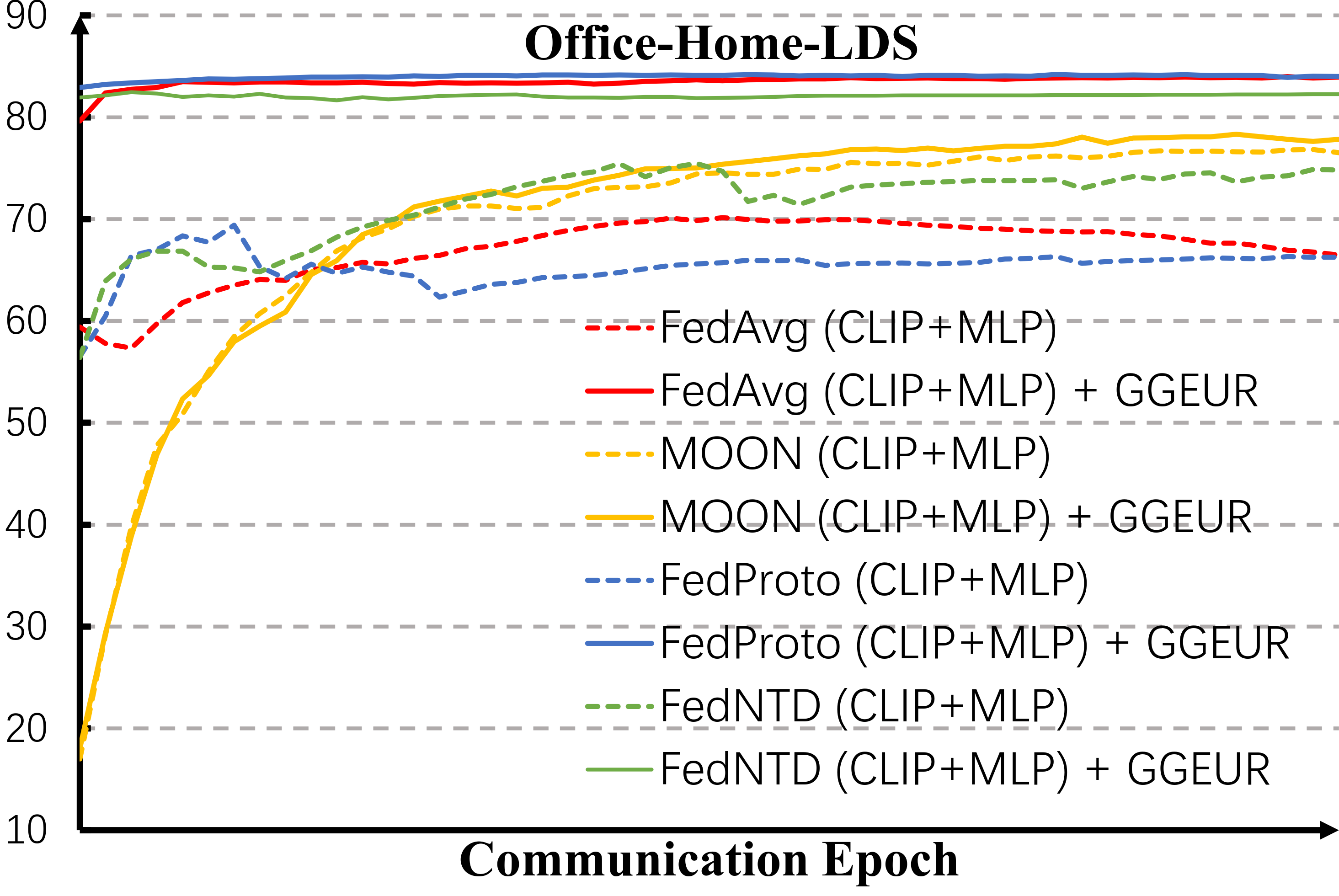}
	\end{minipage}
\vskip -0.1in
\caption{Comparison of convergence in average accuracy with and without integrating GGEUR in the selected FL methods.}
\label{fig8}
\vskip -0.15in
\end{figure}

\subsection{GGEUR Accelerates Convergence}
\label{sec4.5}

In Figure \ref{fig8}, we plot the average accuracy per epoch for various FL methods with and without using GGEUR. It can be observed that our method enhances the convergence speed of the FL methods, resulting in smoother curves.

\section{Conclusion}

This work captures the global geometric shape of the embedding distribution while preserving privacy and utilizes it to simulate an ideal global distribution on clients, bridging the gap between local and global distributions. We first introduce the concept of distributional geometric shapes and address the challenge of obtaining the global geometric shape under privacy constraints. Then, we propose leveraging the global geometric shape to assist clients in augmenting their samples. Extensive experiments demonstrate the effectiveness and compatibility of GGEUR.

\vspace{15px}

\noindent \textbf{Acknowledgement.}
This research is supported by the National Natural Science Foundation of China (Grants 623B2080).

{
    \small
    \bibliographystyle{ieeenat_fullname}
    \bibliography{main}
}

\clearpage

\appendix

\section{Related Work}
\label{rw}
\subsection*{Federated Learning with Data Heterogeneity}

In federated learning, client data often originates from different distributions, typically manifesting as label skew and domain skew. With label skew, the class distributions across clients are significantly imbalanced, while domain skew occurs when feature distributions for the same class vary due to differences in data sources. To address these issues, researchers have proposed methods that can be grouped into client regularization, server-side dynamic aggregation, and federated data augmentation \cite{fl3}.

Client regularization primarily focuses on adjusting local optimization objectives so that local models align more closely with the direction of the global model, reducing distributional shifts across clients \cite{lee2022preservation,fedseg,adcol,dafkd,fedmlb,fedcg,fedrs,fedalign,fedsam,fedlc,feddecorr}. Methods such as FedProx \cite{li2} and SCAFFOLD \cite{SCAFFOLD} introduce additional regularization terms to minimize the discrepancy between local and global models, thereby improving convergence speed and accuracy. MOON \cite{moon} leverages contrastive learning to align feature spaces across clients, addressing both label and domain skew. FPL \cite{huang3} supervises the learning of local class prototypes by aggregating and sharing class prototypes across clients. However, involving a global model in the local optimization process deeply enlarges the local computation cost and linearly increases with the parameter scale.

Server-side dynamic aggregation methods optimize the global model by adaptively adjusting client weights \cite{fedbe,fedgkt,fedavg}. FedOPT \cite{fedopt} and Elastic \cite{chen2023elastic} use dynamic aggregation weights based on client model updates, enhancing the global model’s generalization in heterogeneous data settings. Additionally, methods like FedDF \cite{feddf} and FCCL \cite{huang4} incorporate knowledge distillation modules on the server side, combined with auxiliary datasets to improve the adaptability of aggregation, making these approaches suitable for broader cross-client data distributions. However, these methods often require additional proxy datasets to support model adjustments, which is beneficial in scenarios with significant distributional differences across clients.

Federated data augmentation aim to bridge the gap between local data distributions and the ideal global distribution by generating more diverse data samples on clients \cite{hao2021towards,shin2020xor,ma2023feature}. For example, FedMix \cite{yoon2021fedmix} and FEDGEN \cite{zhu2021data} use MixUp and its variants to augment client data, thus mitigating label skew. However, due to the lack of knowledge-based guidance, these methods largely rely on the diversity of local data. FedFA \cite{zhou2023fedfa} assumes local data follows a Gaussian distribution and generates new samples centered on class prototypes. Nevertheless, the Gaussian assumption is overly idealistic \cite{ma2024geometric}, limiting the ability of generated samples to adequately reduce the discrepancy between local and global distributions. \textbf{Compared to} the Gaussian assumption, the geometric shapes proposed in this work provide a more accurate description of embedding distributions. GGEUR estimates the geometric shape of the global distribution without compromising privacy and leverages it to guide data augmentation on clients. 

\section{Dataset}
\label{Dataset}

\textbf{Label Skew Datasets.} We evaluate our method on three single-domain image classification tasks.
\begin{itemize}[]
    \item \textbf{Cifar-10} \cite{cifar} contains $10$ classes, with $50,000$ images for training and $10,000$ images for validation.
    \item \textbf{Cifar-100} \cite{cifar} covers $100$ classes, with $50,000$ training images and $10,000$ validation images.
\item \textbf{Tiny-ImageNet} \cite{tiny} is the subset of ImageNet with $100K$ images of size $64 \times 64$ with $200$ classes scale.
\end{itemize}
Consistent with recent benchmarks \cite{fl3}, we set up $10$ clients for each task. To simulate label skew across clients, we use a Dirichlet distribution, $Dir(\beta)$, where the parameter $\beta > 0$ controls the degree of label skew (i.e., class imbalance). When $\beta$ takes a smaller value, the local distributions generated on each client become more skewed, showing greater divergence from the overall distribution.

\noindent \textbf{Domain Skew Datasets.} We evaluated our method on the multi-domain image classification dataset Digits and conducted analyses on Office-Caltech and PACS.
\begin{itemize}[]
    \item \textbf{Digits} \cite{MNIST_IEEE98,USPS_PAMI94,svhn_NeurIPS11,syn} includes four domains: MNIST, USPS, SVHN and SYN, each with $10$ categories.
    \item \textbf{Office-Caltech} \cite{office_caltech} includes four domains: Caltech, Amazon, Webcam, and DSLR, each with $10$ categories.
\item \textbf{PACS} \cite{pacs} includes four domains: Photo (P) with $1, 670$ images, Art Painting (AP) with $2,048$ images, Cartoon (Ct) with $2,344$ images and Sketch (Sk) with $3,929$ images. Each domain holds seven categories.
\end{itemize}
Consistent with recent benchmarks, in domain skew experiments, each domain is assigned to a separate client, with each client focusing on data from a specific domain. For Digits, each client is allocated $10\%$ of the data from its respective domain. For Office-Caltech and PACS, each client is allocated $30\%$ of the data from its corresponding domain.

\begin{figure*}[t]
\vskip -0.05in
\centering
	\begin{minipage}{0.33\linewidth}
		\centering
		\includegraphics[width=1\linewidth]{new-0.1_label_distribution_heatmap}
	\end{minipage}
	\begin{minipage}{0.33\linewidth}
		\centering
		\includegraphics[width=1\linewidth]{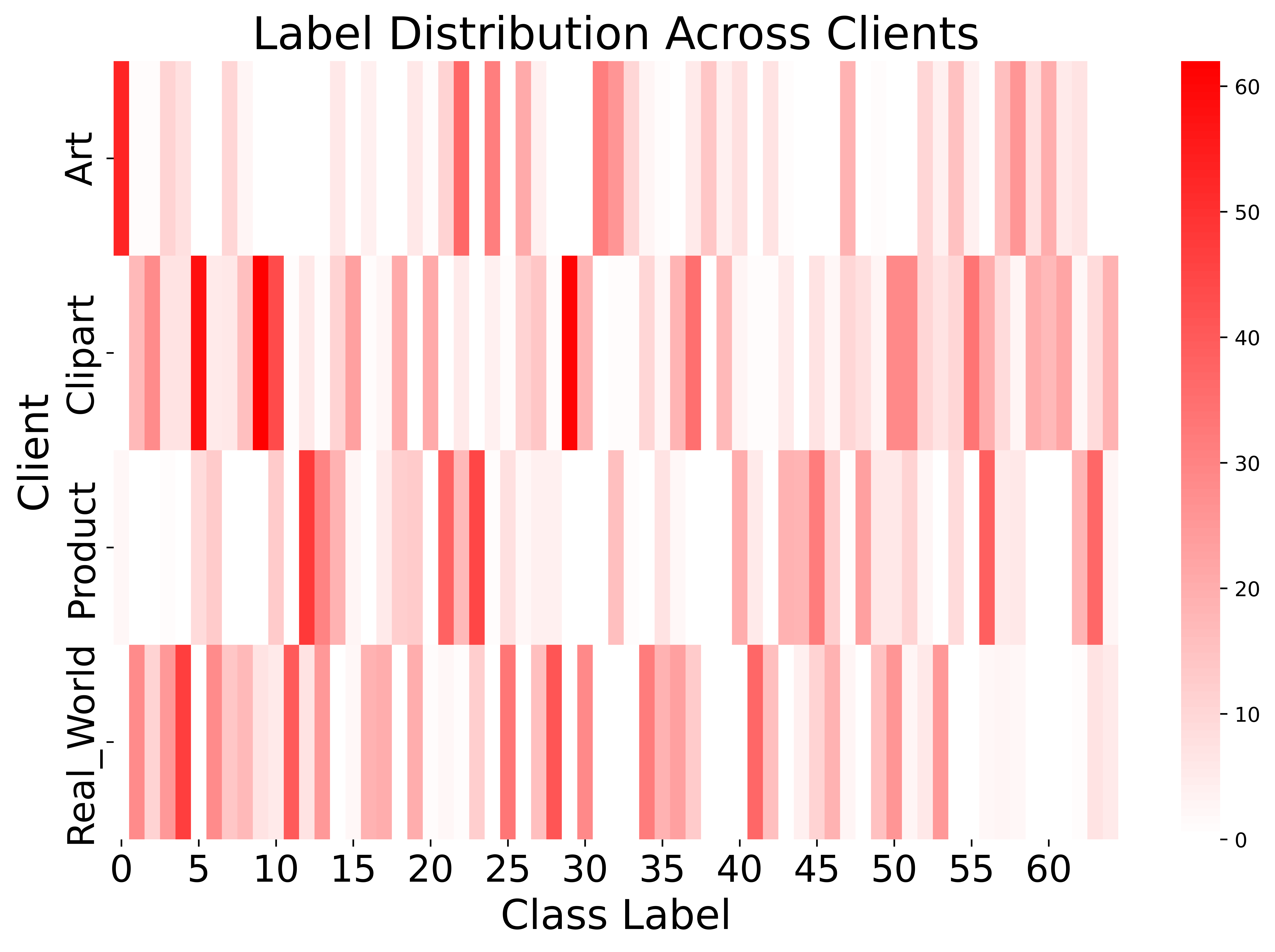}
	\end{minipage}
	\begin{minipage}{0.33\linewidth}
		\centering
		\includegraphics[width=1\linewidth]{new-0.5_label_distribution_heatmap}
	\end{minipage}

	\begin{minipage}{0.33\linewidth}
		\centering
		\includegraphics[width=1\linewidth]{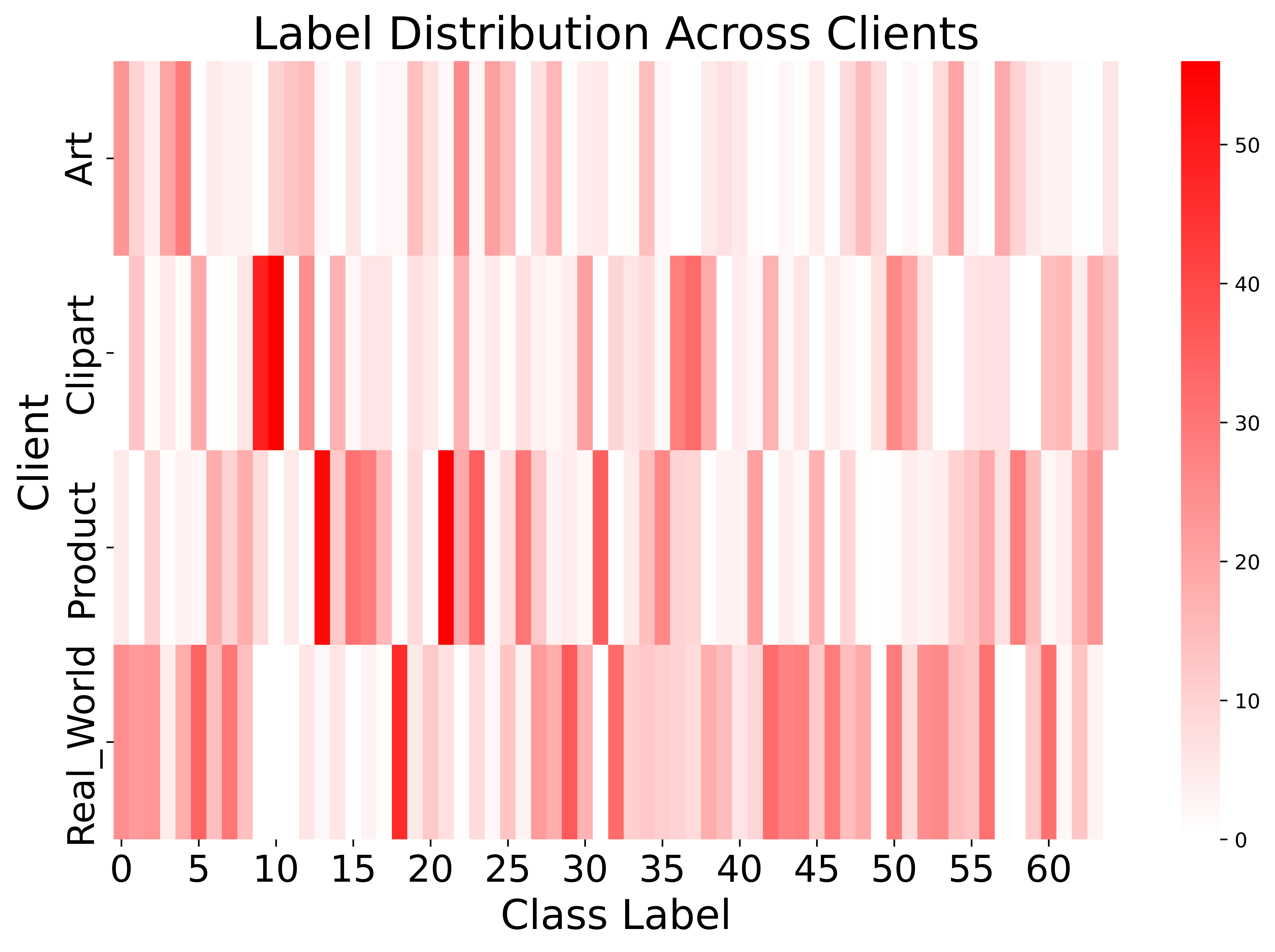}
	\end{minipage}
	\begin{minipage}{0.33\linewidth}
		\centering
		\includegraphics[width=1\linewidth]{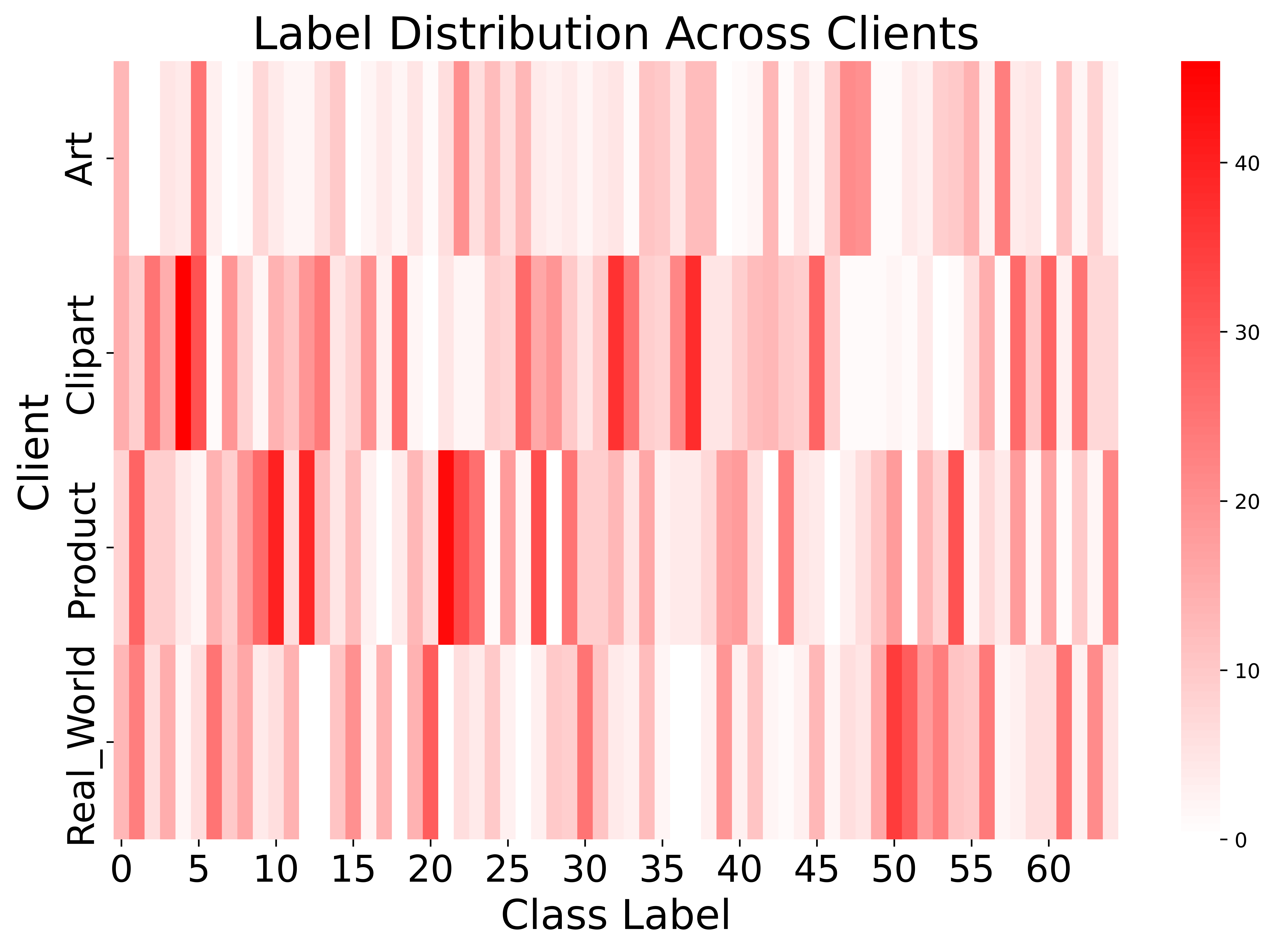}
	\end{minipage}
	\begin{minipage}{0.33\linewidth}
		\centering
		\includegraphics[width=1\linewidth]{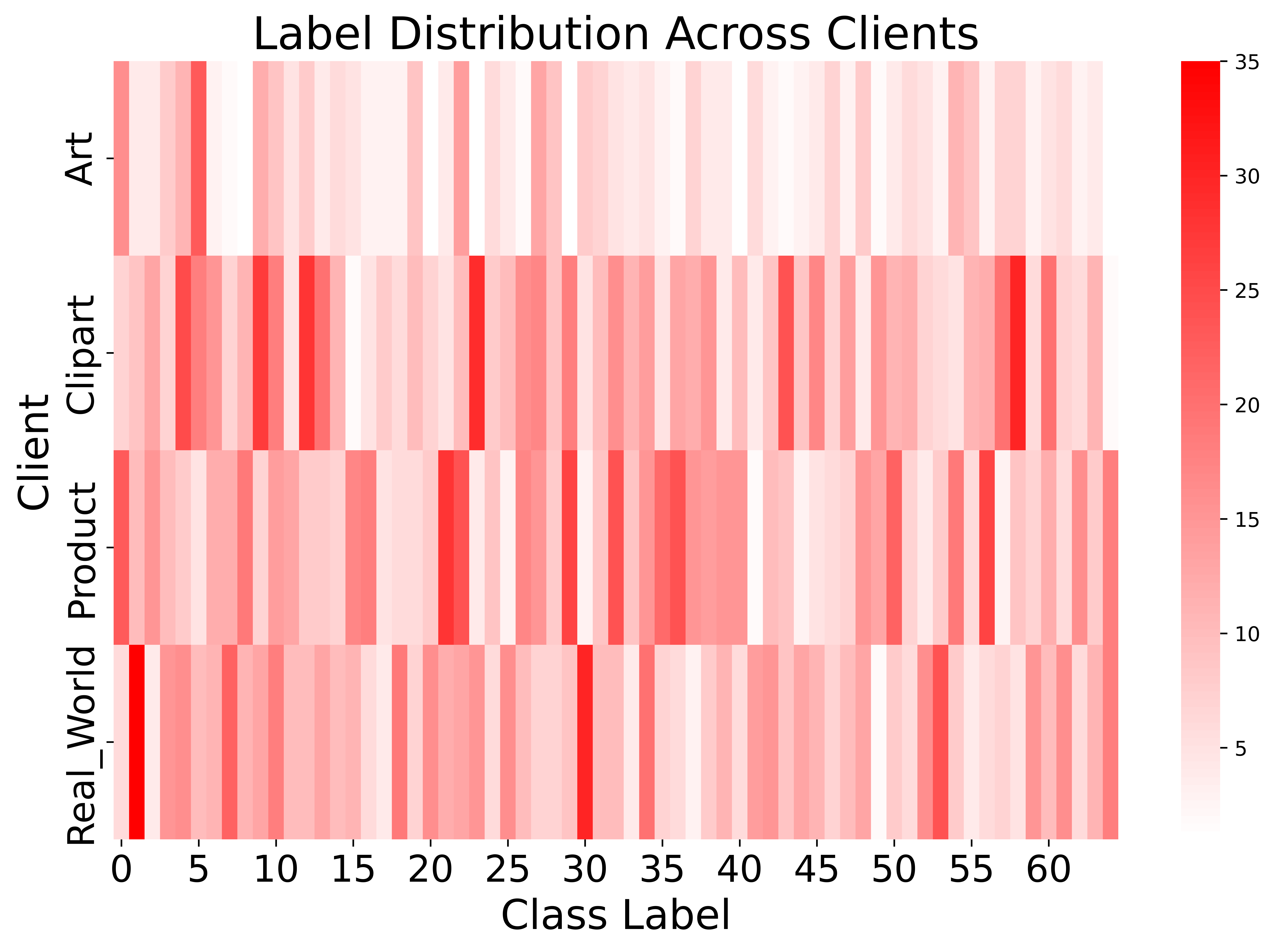}
	\end{minipage}
\vskip -0.1in
\caption{Number of samples per class across four clients when $\beta$ equals $0.1$, $0.3$, $0.5$, $0.7$, $1$, and $5$, with each client holding data from a different domain.}
\label{fig9}
\vskip -0.1in
\end{figure*}

\noindent \textbf{Dataset with Coexisting Label Skew and Domain Skew.}
\textbf{Office-Home} \cite{office_home} includes $4$ domains: Art (A), Clipart (C), Product (P), and Real World (R), each containing $65$ classes. To increase the challenge, we designed a new partitioning method for the multi-domain dataset Office-Home to create a scenario where label skew and domain skew coexist.

In the Office-Home dataset, while ensuring that each client corresponds to a single domain, we first generate a Dirichlet coefficient matrix, where the degree of class imbalance is controlled by $\beta$. For the $65$-class, $4$-domain Office-Home task, we generate a $4\times65$ matrix controlled by $\beta$ (with each column summing to $1$). The four coefficients for each class are then allocated to the four clients, and each client uses its assigned coefficients to determine the number of samples for that class. This setup results in a distribution that incorporates both domain shift (one domain per client) and Dirichlet-based class imbalance, presenting a scenario where the model faces both class distribution and domain differences, creating a more realistic, challenging, and diverse distribution for classification.
We name the newly constructed dataset \textbf{Office-Home-LDS (Label and Domain Skew)}. Figure \ref{fig9} shows the data distribution of Office-Home-LDS with different $\beta$ values. Dataset and Constructor published at: \url{https://huggingface.co/datasets/WeiDai-David/Office-Home-LDS}.

\begin{table}[t]
\centering
\setlength{\tabcolsep}{0.6pt} 
\renewcommand\arraystretch{1}
\caption{Experiments Configuration of different federated scenarios. Image size is operated after the resize operation. $|C|$ denotes the classification scale. $|K|$ denotes the clients number. $E$ is the communication epochs for federation. $B$ is the training batch size.}
\label{table8}
\begin{tabular}{@{}l|c|c|c|c|c|cc@{}}
\toprule
\textbf{Scenario}                       & \textbf{Size} & $|C|$ & \textbf{Network $w$} & \textbf{Rate $\eta$} & $|K|$ & $E$  & $B$  \\ \midrule
\multicolumn{8}{l}{\textbf{Label Skew Setting § \textcolor{red}{5.2}}}                                                             \\ \midrule
Cifar-10                                & 224            & 10    & CLIP (ViT-B/16)            & 1e-2                 & 10    & 100  & 64   \\
Cifar-100                               & 224            & 100   & CLIP (ViT-B/16)            & 1e-1                 & 10    & 100  & 64   \\
Tiny-ImageNet                           & 224            & 200   &CLIP (ViT-B/16)            & 1e-2                 & 10    & 100  & 64   \\ \midrule
\multicolumn{8}{l}{\textbf{Domain Skew § \textcolor{red}{5.3 and 5.4}}}                                       \\ \midrule
Digits                                  & 224            & 10    & CLIP (ViT-B/16)            & 1e-2                 & 4   & 50   & 16   \\
Office Caltech                          & 224           & 10    &CLIP (ViT-B/16)           & 1e-3                 & 4   & 50   & 16   \\
PACS                                    & 224           & 7     &CLIP (ViT-B/16)         & 1e-3                 & 4  & 50   & 16   \\
Office-Home                             & 224           & 65    &CLIP (ViT-B/16)            & 1e-3                 & 4   & 50   & 16   \\ \bottomrule
\end{tabular}
\end{table}

\begin{table}[t]
\centering
\caption{Hyper-parameters chosen for different methods. Hyper-parameters in different methodologies may share the same notation but represent distinct meanings.}
\label{federated_config}
\begin{tabular}{@{}l|l|c@{}}
\toprule
\textbf{Methods} & \textbf{Hyper-Parameter}                  & \textbf{Parameter value} \\ \midrule
SCAFFOLD         & Global learning rate $lr$      & 0.25            \\ \midrule
\multirow{2}{*}{MOON} & Contrastive temp $\tau$    & 0.5             \\ 
                  & Proximal weight $\mu$          & 1.0             \\ \midrule
FedDyn           & Proximal weight $\alpha$       & 0.5             \\ \midrule
FedOPT           & Global learning rate $\eta_g$  & 0.5             \\ \midrule
FedProto         & Proximal weight $\lambda$      & 2               \\ \midrule
\multirow{2}{*}{FedNTD} & Distill temp $\tau$     & 1               \\ 
                  & Reg weight $\beta$             & 1               \\ \bottomrule
\end{tabular}
\end{table}

\section{Implementation Details}
\label{Implementation Details}

As for the uniform comparison evaluation, we follow \cite{fl3} and conduct the local updating round $U=10$. We use the SGD optimizer for all local updating optimization. The corresponding weight decay is $1 \times 10^{-5}$ and momentum is $0.9$. The learning rate $\eta$ and communication epoch $E$ are different in various scenarios, as shown in Table \ref{table8}. Notably, the communication epoch is set according to when all federated approaches have little or no accuracy gain with more communication epochs. The local training batch size is $B = 64$. Furthermore, the Table \ref{federated_config} plots the chosen hyper-parameter for different methods.

\section{Privacy Constraints}
\label{Privacy Constraints}
In our approach, the server only sends the eigenvectors and eigenvalues of the global covariance matrix back to clients, without sharing raw data or local covariance matrices. We demonstrate below that this information is insufficient for reconstructing any client’s original data.

\textbf{(1) Eigenvectors and Eigenvalues Do Not Contain Raw Data.}  
The eigendecomposition provides only the geometric structure of the data distribution, without encoding individual sample details. Even if a client obtains eigenvectors and eigenvalues, reconstructing the original data is an \textbf{ill-posed problem}, as it admits infinitely many solutions.

\textbf{(2) Low-Rank Property Prevents Data Reconstruction.}  
The covariance matrix is typically \textbf{low-rank}, meaning: $\text{rank}(\Sigma_i) \ll d$, where $d$ is the original data dimension. This implies that even with full knowledge of eigenvectors and eigenvalues, clients can only access principal directions of the data and not its full details.

\textbf{(3) Aggregation Prevents Isolation of Individual Client Contributions.}  
The global covariance matrix is an aggregate of all clients’ local covariance matrices: $\Sigma_i = \sum_{k=1}^{K} \frac{n_k^i}{N_i} \Sigma_k^i + \sum_{k=1}^{K} \frac{n_k^i}{N_i} (\mu_k^i - \mu_i)(\mu_k^i - \mu_i)^T.$
Since each client’s contribution is mixed through weighted averaging:
I. No single client can \textbf{isolate another client’s contribution} from Geometric Knowledge.
II. Even if a client’s data is removed, the impact on $\Sigma_i$ is distributed across all eigenvectors, making individual influences indistinguishable.

\textbf{(4) Existing Literature Supports the Privacy of Covariance Matrices.}  
Prior works confirm that sharing higher-order statistics (covariance matrices) poses lower privacy risks than sharing model gradients (Melis et al.).

\section{Large-Scale Client}
\label{Large-Scale Client}
 In practical federated learning settings, the number of participating clients can significantly affect model performance due to increased data heterogeneity. To further evaluate the performance of our proposed method under larger-scale federated learning scenarios, we conducted additional experiments with an increased number of clients. Specifically, we conducted experiments on the label-skewed dataset CIFAR-10 with 100, 300, and 500 clients. As shown in Table \ref{tab11}, the results demonstrate that GGEUR remains robust and continues to enhance the performance of FedAvg (CLIP+MLP).
\begin{table}[h]
\vskip -0.15in
\small
\setlength{\abovecaptionskip}{0cm}
\centering
\setlength{\tabcolsep}{2pt} 
\renewcommand\arraystretch{0.8}
\caption{Number of Clients $K$ Impact on Performance.}
\scalebox{1}{
\begin{tabular}{r||ccc}
\hline\thickhline
\rowcolor{lightgray}
& \multicolumn{3}{c}{CIFAR-10 ($\beta = 0.1$)}  \\
\cline{2-4}
\rowcolor{lightgray}
\multirow{-2}{*}{Methods} 
 &$K=100$  &$K=300$  &$K=500$    \\
\hline\hline

FedAvg (CLIP+MLP) & 87.89 & 84.69  & 82.05      \\
+ \textbf{GGEUR} & 93.55 ($+5.66$)  &92.17 ($+7.84$)   &90.43 ($+8.38$)   \\
\bottomrule \hline
\end{tabular}}
\label{tab11}
\vskip -0.15in
\end{table}

\section{Computational Cost}
\label{Computational Cost}
We conducted experiments on the domain-skewed dataset Digits, comparing the training time required to complete the full model for FedAvg (CLIP+MLP), SCAFFOLD (CLIP+MLP), and MOON (CLIP+MLP) before and after applying GGEUR. The results (Tab\ref{tab10}) show that GGEUR introduces almost no additional training time overhead. Specifically, after applying GGEUR, the training time for the three methods increased by only 3.5s, 4.6s, and 3.3s, respectively. 
\vspace{10px}
\begin{table}[h]
\vskip -0.3in
\small
\setlength{\abovecaptionskip}{0cm}
\centering
\setlength{\tabcolsep}{8pt} 
\renewcommand\arraystretch{0.8}
\caption{The average training time (s) per round.}
\scalebox{1}{
\begin{tabular}{r||ccc}
\hline\thickhline
\rowcolor{lightgray}
& \multicolumn{3}{c}{Digits}  \\
\cline{2-4}
\rowcolor{lightgray}
\multirow{-2}{*}{Methods} 
 &FedAvg  &SCAFFOLD  &MOON    \\
\hline\hline

CLIP+MLP & 28.2 & 54.5  & 32.3      \\
+ \textbf{GGEUR} & 31.7 (+3.5)  &59.1 (+4.6)   &35.6 (+3.3)   \\
\bottomrule \hline
\end{tabular}}
\label{tab10}
\vskip -0.25in
\end{table}

\setlength{\textfloatsep}{18pt}
\begin{algorithm}[t]
\caption{GGEUR (Multi-Domain Scenario)}
\label{alg2}
\begin{algorithmic}[1]
\Require $X_k^i = [X_k^{(i,1)}, \dots, X_k^{(i, n_k^i)}] \in \mathbb{R}^{p \times n_k^i}$: Sample set of class $i$ at client $k$, 
$GD_i = \{\xi_i^1, \dots, \xi_i^p, \lambda_i^1, \dots, \lambda_i^p\}$: Shared geometric shape (eigenvectors and eigenvalues) of class $i$, 
$\{\mu_{k'}^i\}$: Prototypes (means) of class $i$ from other domains, 
$N$: Number of new samples to generate per original sample in Step 1,
$M$: Number of samples to generate per prototype in Step 2.
\Ensure $X_{\text{new}}^i$: Augmented sample set of class $i$ at client $k$

\State $X_{\text{new}}^i \gets \emptyset$ \textcolor{cvprblue}{\Comment{Initialize augmented sample set}}

\textcolor{cvprblue}{\Comment{Step 1: Local Domain Augmentation}}
\For {$j = 1$ to $n_k^i$}
    \State $X_{\text{new}}^i \gets X_{\text{new}}^i \cup \textsc{GGEUR}(X_k^{(i,j)}, GD_i, N)$
\EndFor

\textcolor{cvprblue}{\Comment{Step 2: Cross-Domain Simulation}}
\For {each prototype $\mu_{k'}^i$ from other domains} 
    \State $X_{\text{new}}^i \gets X_{\text{new}}^i \cup \textsc{GGEUR}(\mu_{k'}^i, GD_i, M)$
\EndFor

\State \textbf{return} $X_{\text{new}}^i$
\end{algorithmic}
\end{algorithm}

\end{document}